\pdfoutput=1
\documentclass{article}

\usepackage{glyphbench_preprint,times}
\usepackage{amsmath,amsfonts,bm}

\def\eqref#1{equation~\ref{#1}}
\def\1{\bm{1}}

\DeclareMathAlphabet{\mathsfit}{\encodingdefault}{\sfdefault}{m}{sl}
\SetMathAlphabet{\mathsfit}{bold}{\encodingdefault}{\sfdefault}{bx}{n}

\usepackage{booktabs}
\usepackage{graphicx}
\usepackage{microtype}
\usepackage{placeins}
\usepackage{xcolor}
\usepackage{hyperref}
\usepackage{url}


\newcommand{\NumStandardTasks}{303}
\newcommand{\NumExtendedTasks}{59}
\newcommand{\NumAllTasks}{362}
\newcommand{\BaselineModel}{Qwen3.5-4B}
\newcommand{\ComparisonModel}{Qwen3.8-27B}
\newcommand{\NumEvalTasks}{303}
\newcommand{\NumEvalEpisodesPerModel}{7575}
\newcommand{\QwenMeanGain}{0.436}
\newcommand{\NumSingleTaskRuns}{25}
\newcommand{\NumMultitaskTrainTasks}{100}

\newcommand{\HeroStep}{250}
\newcommand{\HeroEvalInitial}{0.012}

\newcommand{\HeroEvalEarly}{0.319}

\definecolor{GlyphGreen}{HTML}{087843}

\hypersetup{
  colorlinks=true,
  linkcolor=GlyphGreen,
  citecolor=GlyphGreen,
  urlcolor=GlyphGreen,
  pdftitle={GlyphBench: A Playground for Language-Model Reinforcement Learning},
  pdfauthor={Roger Creus Castanyer; Marc-Alexandre C\^ot\'e; Matthew James Sargent; Augustine N. Mavor-Parker; Glen Berseth; Pablo Samuel Castro}
}

\title{GlyphBench: A Playground for\\Language-Model Reinforcement Learning}

\newcommand{\affillogo}[2][0.95em]{%
  \raisebox{-0.12em}{\includegraphics[height=#1]{#2}}%
}
\newcommand{\udemmark}{\affillogo{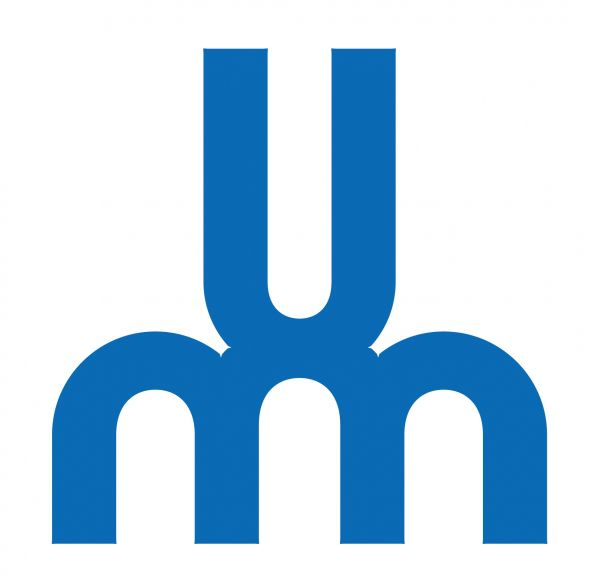}}
\newcommand{\milamark}{\affillogo[1.1em]{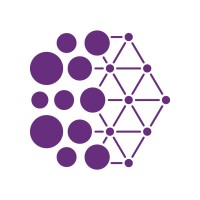}}
\newcommand{\vmaxmark}{\affillogo[0.82em]{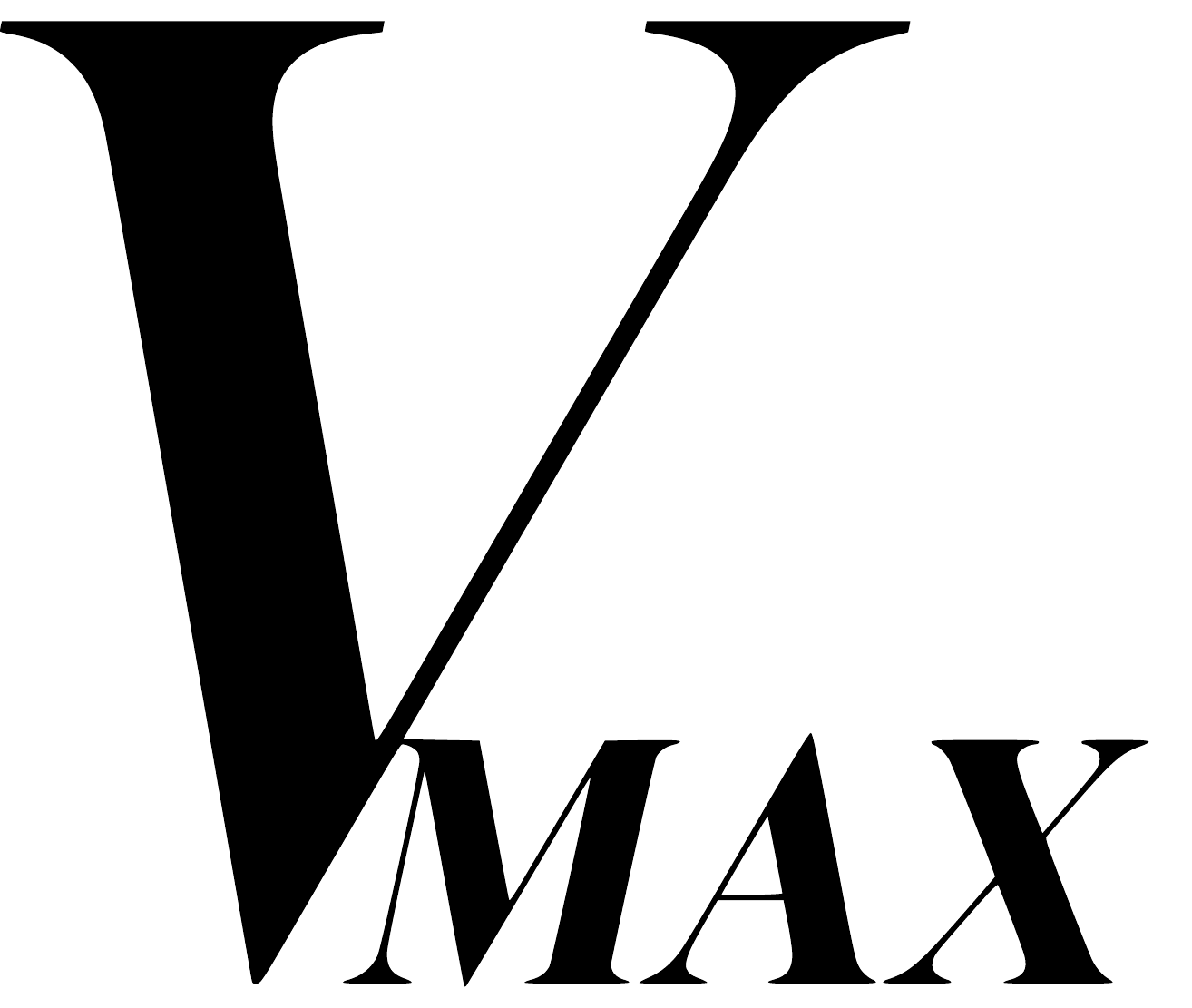}}

\author{%
Roger Creus Castanyer\,\milamark\,\udemmark\,\vmaxmark\quad
Marc-Alexandre C\^ot\'e\,\milamark\quad
Matthew James Sargent\,\vmaxmark\\
\textbf{Augustine N. Mavor-Parker\,\vmaxmark\quad
Glen Berseth\,\udemmark\,\milamark\quad
Pablo Samuel Castro\,\udemmark\,\milamark}\\[0.6em]
\milamark\,Mila - Quebec Artificial Intelligence Institute\\[0.2em]
\udemmark\,Universit\'e de Montr\'eal\\[0.2em]
\vmaxmark\,Vmax
}

\begin{document}

\maketitle

\begin{abstract}
We introduce \textbf{GlyphBench}, an environment suite for reinforcement learning (RL) post-training of language-model agents, with over 360 tasks spanning diverse games. GlyphBench renders spatial observations as two-dimensional Unicode grids and connects training, evaluation, and trajectory replay through a unified interface designed to support efficient and reproducible research. We use GlyphBench to study how observation interfaces, reasoning effort, and agent harnesses affect performance, and how RL configurations shape learning dynamics. Our results show that glyph observations outperform native text and pixels in our Craftax experiments, with further gains on several BALROG environments. RL on 100 GlyphBench tasks improves Qwen3.5-4B on held-out Reasoning Gym problems, reaching 63.48\% accuracy and outperforming the base model, a math-trained baseline, and a code-trained baseline. These experiments provide empirical evidence that reasoning gains from gameplay can yield stronger transfer than math or code. Together, these results highlight GlyphBench's value as a testbed for systematic research on how language-model agents learn, interact, and generalize.

\end{abstract}
\bibliographystyle{glyphbench_preprint}
% Shared manuscript content.
\section{Introduction}
\label{sec:introduction}

Reinforcement learning (RL) has improved the performance of language models on problems with verifiable rewards, such as mathematics and code \citep{guo2025deepseek}. Tasks that are interactive by nature are particularly challenging: models must choose actions that shape subsequent observations, act with incomplete information, and account for consequences that may only become apparent later. For example, a software-engineering agent edits code, runs tests, and uses the resulting feedback to guide further decisions \citep{golubev2025multiturn}. To improve agents' reasoning, we need to understand how they use the information available at each decision, how they learn from environmental feedback, and whether the capabilities they acquire extend to new problems. Studying these questions requires diverse interactive tasks that researchers can use for training and evaluate under controlled conditions.

Games have been a central testbed for deep RL over the past decade, exposing different learning challenges through repeatable interactions. Benchmarks such as the ALE \citep{bellemare2013arcade} helped establish deep RL \citep{mnih2015human}, while Procgen \citep{cobbe2020leveraging}, MiniHack \citep{samvelyan2021minihack}, and Craftax \citep{matthews2024craftax} expose distinct challenges in generalization, exploration, and planning. Games have also driven landmark results in Go and StarCraft II \citep{silver2017mastering,vinyals2019grandmaster}, which have sparked further advances in the field. Collectively, these benchmarks support both targeted studies of individual capabilities and experiments in which several capabilities must work together.

For language-model post-training, however, differences in observation and action interfaces complicate comparisons across environments: performance depends on how information is presented as well as on the decisions a task requires \citep{paglieri2024balrog}. Recent training studies use settings as different as BabyAI-Text \citep{carta2023glam}, symbolic multi-turn games \citep{wang2025ragen}, and visual arcade games \citep{xie2025vigal}. BALROG and lmgame-Bench bring several games together through shared interfaces \citep{paglieri2024balrog,hu2025lmgame}, but game collections, observation formats, and training setups still differ across studies. Making these well-studied environments accessible through a common text interface that preserves spatial layout offers a practical way to study learning and generalization across games within a shared experimental framework.

\begin{figure}[!htb]
  \centering
  \includegraphics[width=\linewidth]{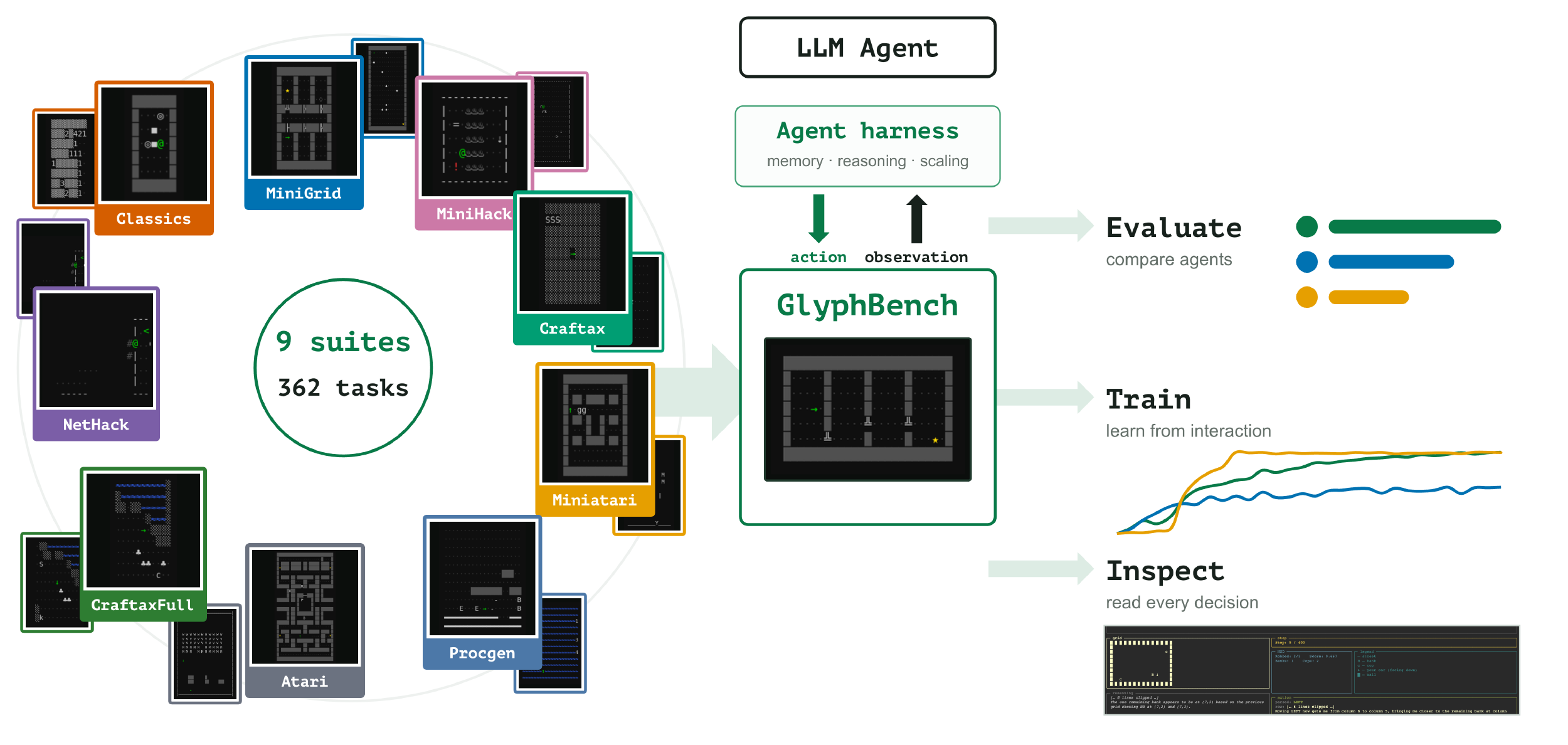}
  \caption{GlyphBench connects nine game families through spatial text observations and named actions. A shared task interface supports policy evaluation, RL training, and trajectory replay. Six families form the standard track; three provide extended tasks.}
  \label{fig:overview}
\end{figure}

We introduce \textbf{GlyphBench} to study these questions across a diverse collection of interactive environments. GlyphBench brings together puzzles, arcade games, and established RL benchmarks through a common text interface: agents observe two-dimensional Unicode grids and interact through named actions. These grids preserve the relative positions of objects in a form that language models can consume directly and researchers can inspect. Researchers can train and evaluate policies across these environments, replay trajectories, and test changes to training or the agent harness (Figure~\ref{fig:overview}). GlyphBench supports this cycle of measuring performance, diagnosing behavior, and testing improvements across diverse tasks.

We first examine how models act before additional training, comparing task performance, episode length, and token use across models of different sizes and capabilities. Our experiments in Craftax \citep{matthews2024craftax} show the benefit of spatial glyph observations: they achieve higher mean reward than native text and pixels for all three models we evaluate. We separately examine how harness design affects the use of additional reasoning. Single-task policies improve across all six standard suites, revealing a range of learning challenges alongside differences in existing model capabilities.

To illustrate how researchers can investigate training dynamics, we vary learning rate and rollout group size in Snake under a fixed interaction budget. Smaller rollout groups allow more updates, but the resulting gains depend strongly on learning rate. This experiment illustrates GlyphBench's use as a tool for studying how training choices interact.

Finally, we study joint post-training and transfer. A single policy trained on 100 GlyphBench tasks improves on held-out environment seeds across the six suites. The trained policy also improves on held-out Reasoning Gym \citep{stojanovski2025reasoninggym} problems, outperforming a math-trained baseline and a code-trained baseline. These results provide empirical evidence that reasoning gains from gameplay can transfer more effectively to this benchmark than those from math or code RL. Together, our evaluations demonstrate GlyphBench's usefulness as a benchmark, a research tool, and a post-training suite, and provide broader insight into the value of games for understanding and improving language models.

\section{Related Work}
\label{sec:related}

\paragraph{Language-model agents in games.}
Applying language models to game environments introduces a fundamental design question: how should pretrained knowledge be translated into effective actions? SPRING addresses this through structured reasoning over knowledge extracted from the environment \citep{wu2023spring}, while Voyager generates, refines, and reuses executable skills in Minecraft \citep{wang2023voyager}. Both illustrate how an agent's reasoning procedure and supporting components can improve gameplay without updating the underlying model's parameters. Broader benchmarks examine these design choices across environments. BALROG evaluates language-model and vision-language agents on established RL tasks and identifies limitations in handling complex environments and visual observations \citep{paglieri2024balrog}. lmgame-Bench studies how perception, memory, and prompt design affect the reliability of game-based evaluation \citep{hu2025lmgame}. Agentick extends comparisons to RL, language-model, and vision-language agents through common tasks with multiple observation modalities \citep{castanyer2026agentick}. These studies show why comparisons must account for the interaction interface alongside the underlying model, motivating our experiments on harness design and inference-time reasoning.

\paragraph{Learning through interaction and transfer.}
Learning from interaction raises two further questions: how to train policies effectively, and whether the resulting improvements extend beyond the training tasks. GLAM investigates online RL for grounding pretrained language-model policies in textual environments with spatial and navigation tasks \citep{carta2023glam}. RAGEN studies training instability and the effects of rollout diversity and freshness in multi-turn RL \citep{wang2025ragen}, while AgentGym-RL develops a framework for training across interactive environments and a strategy that progressively increases interaction horizons \citep{xi2025agentgymrl}. These works make the collection and optimization of interaction trajectories central to the study of agent learning. Evidence for transfer from games also highlights the importance of the evaluation domain. The initial lmgame-Bench study reports that RL on Sokoban or Tetris improves performance on other games, planning problems, and WebShop \citep{hu2025lmgame}. ViGaL reports that RL on visual arcade games improves multimodal reasoning on benchmarks including MathVista and MMMU \citep{xie2025vigal}. These findings motivate investigating which capabilities transfer under different training conditions. We use GlyphBench to study optimization under a fixed interaction budget and multitask training across diverse spatial text games, evaluating both performance on held-out game seeds and transfer to held-out Reasoning Gym problems \citep{stojanovski2025reasoninggym}. Appendix~\ref{app:related} expands these connections.
\section{GlyphBench}
\label{sec:benchmark}

GlyphBench provides over 360 tasks across nine game families through a common framework for language-model training, evaluation, and trajectory replay. Researchers can study learning on individual tasks, train policies across task mixtures, and inspect the decisions behind aggregate performance. Glyph observations and named, task-specific actions provide a consistent interaction protocol across the suite. Standard tasks bound episode lengths and returns to support repeated RL experiments, while extended tasks support longer episodes of exploration and progression.

\subsection{Tasks, observations, and agent harnesses}
\label{sec:contract}

Each task $e$ is an episodic, partially observed decision process with transition rule $P_e$, observation rule $O_e$, and rewards $r_t$. A language model with parameters $\theta$ and a harness $\eta$ induce a policy $\pi_{\theta,\eta}(a_t\mid h_t)$ over actions given interaction history $h_t$. The harness constructs prompts, retains history or memory, and parses model outputs. For task distribution $\mathcal{D}$, training seeks to maximize
\begin{equation}
  J_{\mathcal{D}}(\theta;\eta)
  = \mathbb{E}_{e\sim\mathcal{D},\,\tau\sim(\pi_{\theta,\eta},e)}
    \bigl[R_e(\tau)\bigr],
  \qquad R_e(\tau)=\sum_{t=0}^{T-1}r_t,
  \label{eq:objective}
\end{equation}
where $\tau$ is an episode of length $T$. This notation distinguishes the variables studied below: the observation rule and harness determine the agent's inputs, RL updates $\theta$, and the training mixture determines $\mathcal{D}$.

\begin{figure}[!htb]
  \centering
  \includegraphics[width=0.88\linewidth]{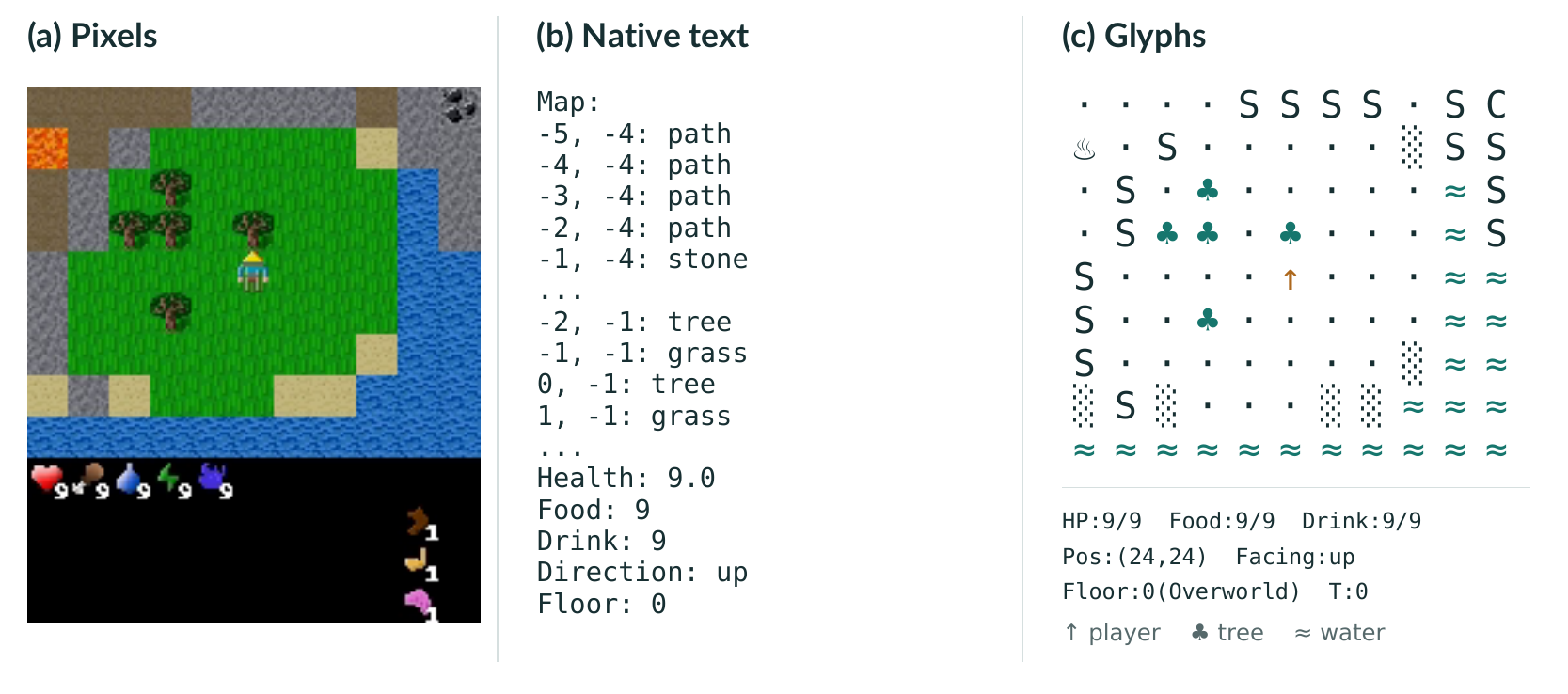}
  \caption{The same Craftax state in pixels, Craftax's original text renderer (native text), and GlyphBench's glyph format. Native text is a comparison interface, not part of GlyphBench. We show excerpts here and both complete text observations in Figure~\ref{fig:craftax-full-observations}. Glyph colors aid visualization only.}
  \label{fig:craftax-interface-modalities}
\end{figure}

We place spatial information in a two-dimensional Unicode grid, identify symbols with a legend, and use a heads-up display (HUD) for quantities such as health, inventory, or velocity. Actions use a task-specific discrete vocabulary. This representation makes adjacency and layout explicit in text (Figure~\ref{fig:craftax-interface-modalities}); Appendix~\ref{app:environment} shows the full interface in replay.

\subsection{Standard and extended tasks}

We select game families with an established role in RL research on control, procedural generalization, navigation and memory, and survival with staged progression (Appendix Table~\ref{tab:suites}). Using these familiar challenges gives researchers a basis for interpreting language-model behavior and choosing tasks that test a particular hypothesis.

We reimplement their mechanics from scratch in \NumStandardTasks{} standard tasks and bound episode lengths and cumulative returns by design: $T\leq512$ and $R_e(\tau)\in[-1,1]$. The episode cap limits the cost of collecting each training example, while the shared return range makes it easier to combine tasks and compare progress across a mixture. The \NumExtendedTasks{} extended tasks support longer episodes through arcade reimplementations and wrappers around the existing full Craftax and NetHack engines. We use standard tasks for RL training and the full Craftax game to study agents that must explore, gather resources, and prepare for challenges across nine floors. Appendix~\ref{app:craftax-interface} illustrates these floors, and Appendix~\ref{app:environment} describes the training interfaces and trajectory replay tools.

\section{Experimental Setup}
\label{sec:protocol}

We evaluate GlyphBench as a benchmark for agents, a tool for research on learning dynamics, and a suite for post-training. We first compare existing models and train individual task policies to examine the capabilities and learning challenges captured by the benchmark. We then use Snake to illustrate how researchers can study the interaction between optimization choices. Finally, we train a policy on a task mixture and evaluate it on external reasoning problems to test whether learning through gameplay transfers beyond the training environments. Training uses our Prime-RL integration \citep{prime2026rl}; Appendix~\ref{app:training} specifies the objectives and configurations.

\paragraph{Benchmarking agents.}
We evaluate \BaselineModel{} and \ComparisonModel{} on all \NumEvalTasks{} standard tasks with 25 matched seeds per task (\NumEvalEpisodesPerModel{} episodes per model; Figure~\ref{fig:paired}). Returns are averaged within tasks and tasks receive equal weight. Episode length and prompt-plus-completion tokens complement return by measuring how much interaction and generation each model uses.

We use the full nine-floor Craftax game to study how agents interpret observations and use information from earlier decisions. We compare two agent harnesses, with examples in Appendix~\ref{app:craftax-harnesses}. The \textit{Basic} harness presents the game instructions and conversation history at each turn. The \textit{Guided Memory} harness instead organizes information into a compact memory of plans, landmarks, and recent actions, and supplies guidance for the current floor. It updates this memory through separate model calls, giving the agent a way to revisit its plans as the game changes. Both harnesses select one native game action per turn and cannot execute scripts.

In Figure~\ref{fig:craftax-systems}, we evaluate GPT-5.6-\{Luna, Terra, Sol\}. We first compare native text, glyphs, and pixels with Guided Memory at high reasoning effort. We then compare the two harnesses using glyphs and vary Sol's reasoning effort to examine how harness design affects the use of additional computation. We report five-episode means with 95\% $t$ intervals, scoring reward as $100R/226$ \citep{craftaxscoreboard2026}. We include one exploratory GPT-6-Astra episode with Guided Memory at maximum effort to illustrate how far a stronger model can progress. Appendix~\ref{app:craftax-interface} describes the Craftax interfaces and harnesses. We also compare original and glyph interfaces for five models in BabyAI, Baba Is AI, Crafter, and MiniHack from BALROG \citep{paglieri2024balrog} (Figure~\ref{fig:balrog-interface}, Appendix~\ref{app:balrog-interfaces}).

We train \NumSingleTaskRuns{} separate \BaselineModel{} policies across all six standard suites to check whether the tasks supply useful reward signals and the common training interface supports improvement. Figure~\ref{fig:single} shows selected learning curves; Appendix Figure~\ref{fig:single-full} shows all runs.

\paragraph{Studying learning dynamics.}
To illustrate how GlyphBench supports studies of learning dynamics, we use Snake to examine how learning rate and rollout group size interact during RL. Across its three difficulty levels, we compare frequent updates from small groups with fewer updates from large groups under the same interaction budget (Figure~\ref{fig:snake}). We cross five group sizes (32--512) with four learning rates ($10^{-6}$--$10^{-5}$), yielding 60 conditions. Each update uses one group, so 6,144 rollouts permit $6{,}144/G$ updates at group size $G$. In Appendices~\ref{app:snake-collapse} and~\ref{app:snake-long}, we examine higher learning rates and longer rollout budgets.

\paragraph{Post-training and transfer.}
We then study learning across a mixture of tasks over 100 gradient steps with one \BaselineModel{} policy trained on \NumMultitaskTrainTasks{} GlyphBench tasks. We use group size 32, learning rate $5\!\times\!10^{-6}$, and 32 groups per update (1,024 rollouts), with Prime-RL's IPO-style loss. We evaluate on held-out environment seeds every 50 updates (Figure~\ref{fig:multitask}). 

To test whether learning through interaction improves reasoning outside the training environments, we compare the base model and policies obtained after GlyphBench, Math, and Code RL on identical Reasoning Gym problems \citep{stojanovski2025reasoninggym} (Figure~\ref{fig:transfer}). Appendix~\ref{app:baseline-transfer} describes the baseline datasets and evaluations; Appendix~\ref{app:transfer} reports additional evaluations on MathArena and SWE-bench \citep{balunovic2025matharena,jimenez2024swebench,deng2025swebenchpro}.

\section{Results}
\label{sec:results}

Our experiments establish three uses of GlyphBench. As a benchmark, it distinguishes agent capabilities and exposes varied learning challenges (Section~\ref{sec:observations}). As a research tool, it reveals interactions between optimization choices (Section~\ref{sec:learning}). As a post-training suite, it supports joint learning across games and stronger Reasoning Gym transfer than math or code RL in our main experiment (Section~\ref{sec:transfer}).

\subsection{GlyphBench as a benchmark for agents}
\label{sec:observations}

\begin{figure}[!htb]
  \centering
  \includegraphics[width=\linewidth]{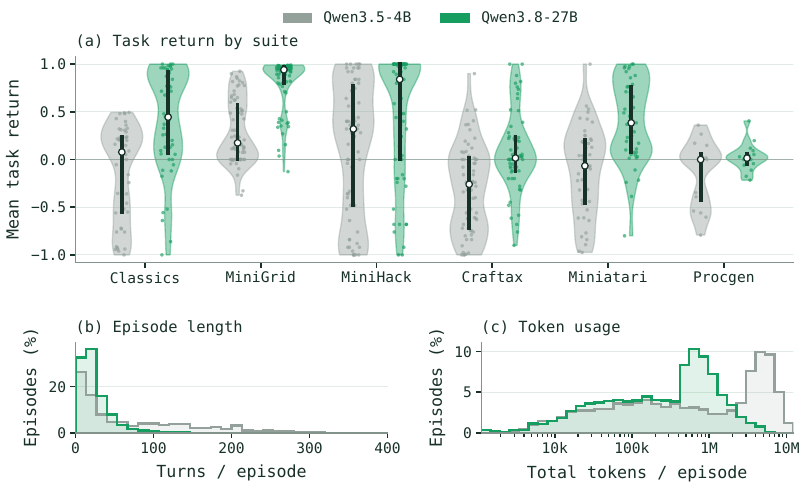}
  \caption{Zero-shot performance on \NumEvalTasks{} tasks with 25 matched seeds each. Panel (a) shows task-mean returns, with interquartile ranges and median dots; (b--c) show episode length and token use over \NumEvalEpisodesPerModel{} episodes per model. Tokens include prompts and completions under each model's tokenizer.}
  \label{fig:paired}
\end{figure}

We first characterize performance before additional RL training on GlyphBench. \ComparisonModel{} improves on 270 of \NumEvalTasks{} tasks, with a mean paired return gain of \QwenMeanGain{}. Return distributions shift upward in every suite, while scores remain spread across tasks within each family (Figure~\ref{fig:paired}). The collection therefore distinguishes the two models while retaining tasks on which the stronger model has room to improve, establishing a reference point for the training experiments that follow.

Higher performance also comes with lower interaction cost: mean episode length falls from 72.8 to 23.9 turns and mean total tokens from 2.15M to 596k. This seemingly counterintuitive result is consistent with a stronger policy reaching goals directly, while a weaker one spends longer repeating unproductive actions or reasoning. Researchers can use GlyphBench's replay interface to examine these behaviors and connect differences in return and token use to individual decisions (Appendix~\ref{app:environment}).

\paragraph{Observation format, harness support, and reasoning effort.}

\begin{figure}[!htb]
  \centering
  \includegraphics[width=\linewidth]{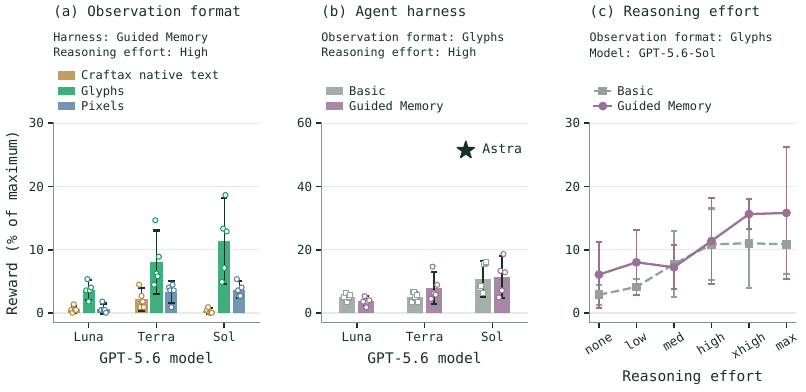}
  \caption{Zero-shot Craftax performance across observation formats (a), harnesses (b), and reasoning efforts (c), as a percentage of maximum reward ($100R/226$). Each panel specifies the settings held fixed. Whiskers show 95\% $t$ intervals around five-episode means; open markers show individual episodes in (a--b). The star marks one exploratory Astra episode with Guided Memory at maximum effort.}
  \label{fig:craftax-systems}
\end{figure}

Our results show that glyph observations give the highest mean reward in full Craftax for Luna, Terra, and Sol with Guided Memory at high reasoning effort (Figure~\ref{fig:craftax-systems}(a)). For Sol, the mean reaches 11.4\% of maximum reward, compared with 3.7\% for pixels and 0.3\% for native text. Glyphs preserve adjacency and spatial layout while naming entities through a legend. This reduces the need to reconstruct a map from coordinate descriptions or identify objects visually, offering a plausible explanation for the advantage. The result also parallels Agentick's findings on spatial text observations \citep{castanyer2026agentick}.

Richer harness support becomes most useful when paired with sufficient reasoning. At high effort, Guided Memory improves Terra and slightly improves Sol, while Luna achieves a higher mean with Basic (panel (b)). Sol's effort sweep makes the interaction clearer (panel (c)): performance with Basic levels off near 11\%, whereas Guided Memory continues improving to 15.8\% at maximum effort. The steeper improvement with Guided Memory suggests that additional reasoning helps Sol use the richer memory and guidance. Harness design therefore affects how much an agent gains from additional inference computation, motivating studies that vary both together \citep{snell2025scaling}.

Astra reaches 51.4\% in the exploratory episode with Guided Memory at maximum reasoning effort, the highest individual return in our Craftax evaluation. This run illustrates how far an agent can progress when a stronger model combines the glyph interface with memory and guidance.

Additionally, we use BALROG \citep{paglieri2024balrog} to examine how glyph rendering affects performance outside the GlyphBench task collection (Figure~\ref{fig:balrog-interface}, Appendix~\ref{app:balrog-interfaces}).

\paragraph{Learning challenges across tasks.}

We next train separate policies to test whether the shared interface and task rewards support learning across all six standard suites. Figure~\ref{fig:single} shows two tasks from each suite, with all \NumSingleTaskRuns{} runs in Appendix~\ref{app:single-task}. Training returns improve across the suites, although the pace and stability of learning differ between tasks. The different learning trajectories show that the suite offers both learnable reward signals and tasks that remain challenging after training. Researchers can therefore study differences in learning speed and stability across tasks through a common interface.

\begin{figure}[!b]
  \centering
  \includegraphics[width=\linewidth]{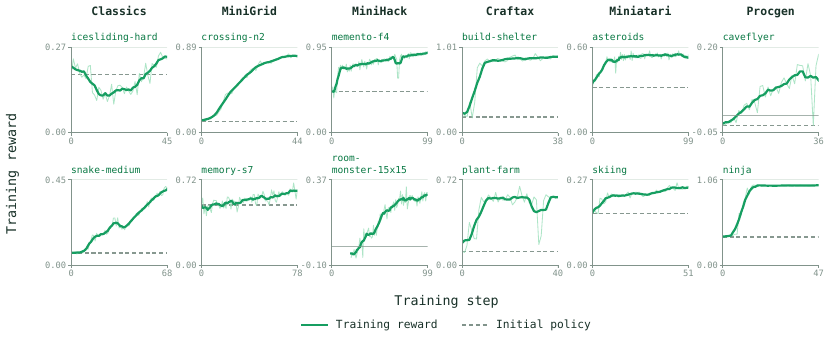}
  \caption{Policies learn on individual tasks across all six suites. Pale traces show training reward; dark traces show centered seven-point means. Dashed lines show mean initial-policy returns. Axes vary by task. Appendix~\ref{app:single-task} explains the selection and shows all \NumSingleTaskRuns{} learning curves.}
  \label{fig:single}
\end{figure}

\subsection{GlyphBench as a tool for research}
\label{sec:learning}

\begin{figure}[!htb]
  \centering
  \includegraphics[width=\linewidth]{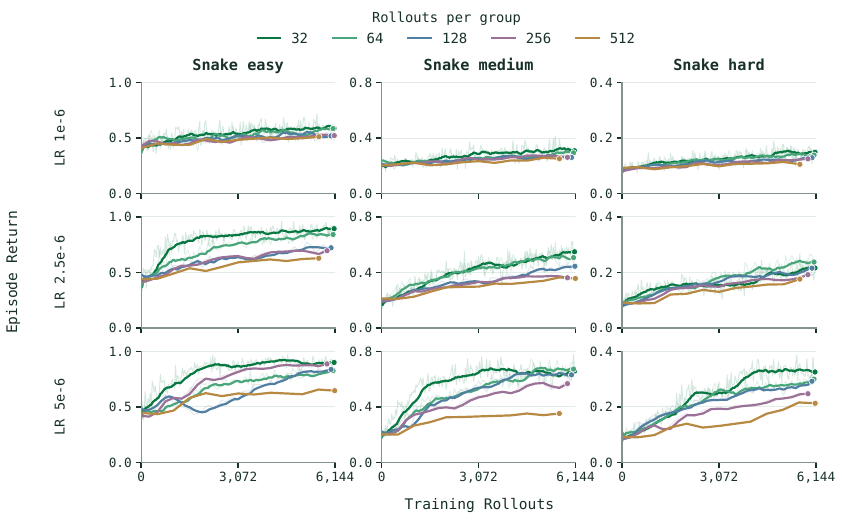}
  \caption{Learning in Snake under a fixed 6,144-rollout budget. The 45 conditions vary learning rate (rows), difficulty (columns), and group size (colors). Pale traces show batch returns; solid curves show trailing 512-rollout means; dots mark final observations. Return scales match within columns. We examine higher learning rates in Appendix~\ref{app:snake-collapse}.}
  \label{fig:snake}
\end{figure}

We use the following experiment to illustrate how GlyphBench supports research on learning dynamics and hyperparameter choices in post-training. Specifically, we examine how a fixed amount of interaction is used for policy improvement. In GlyphBench's Snake environment, we vary rollout group size and learning rate across three difficulty levels while holding the total number of sampled episodes fixed (Figure~\ref{fig:snake}). These choices jointly control the updates made from that experience: smaller groups permit more frequent updates, while learning rate scales each update. A 6,144-rollout budget gives 192 updates at group size 32, compared with 12 at group size 512.

The benefit of frequent updates depends on how much each update changes the policy. At $10^{-6}$, progress is modest and group sizes remain relatively close. At intermediate rates, smaller groups pull ahead: raising the rate to $5\!\times\!10^{-6}$ increases group 32's final return on easy Snake from 0.587 to 0.900, with analogous gains on medium and hard. The precise optimum varies with difficulty, but the common pattern is that learning rate and update frequency must be chosen together.

Larger rates eventually erase these gains: $10^{-5}$ and $10^{-4}$ collapse toward zero (Appendix~\ref{app:snake-collapse}). Conversely, extending the budget to 25,600 rollouts at $10^{-6}$ preserves group 32's advantage over group 512 on all three difficulties (Appendix~\ref{app:snake-long}). Across these Snake tasks, smaller rollout groups are most effective when paired with a suitable learning rate. The experiment illustrates how GlyphBench can be used to study the interaction between optimization choices and task difficulty under a common interaction budget.

\subsection{GlyphBench as a post-training suite}
\label{sec:transfer}

\paragraph{Learning across a task mixture.}

\begin{figure}[!htb]
  \centering
  \includegraphics[width=\linewidth]{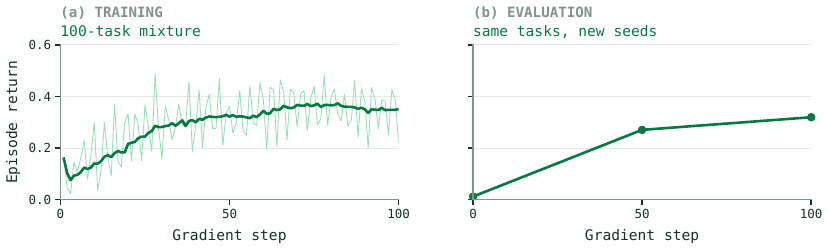}
  \caption{One policy learns across 100 tasks over 100 gradient steps. Panel (a) shows training returns with a trailing nine-step mean; (b) evaluates the same tasks on new environment seeds.}
  \label{fig:multitask}
\end{figure}

The single-task experiments train a separate policy for each task. We now test whether one policy can improve across a mixture of 100 tasks with different rules and reward structures. Mean return on new instances of these tasks rises from \HeroEvalInitial{} to \HeroEvalEarly{} within 100 gradient steps (Figure~\ref{fig:multitask}). These gains show that one policy can learn from rewards across games with different rules, making the mixture a useful source of interactive experience for post-training. Evaluation on new instances tests learning beyond the particular episodes seen during training; the external reasoning evaluation below tests transfer beyond these task types. Appendix~\ref{app:multitask} provides task-level learning curves and evaluations by suite.

To relate these gains to other common domains for language-model RL, we apply the same optimization recipe to a set of 100 math problems and a set of 100 code problems. In Appendix~\ref{app:baseline-transfer}, we show the learning curves for these runs. We next evaluate the policies obtained after training on held-out reasoning problems to compare how their gains transfer beyond their respective training domains.

\paragraph{Transfer beyond gameplay.}

\begin{figure}[!htb]
  \centering
  \includegraphics[width=\linewidth]{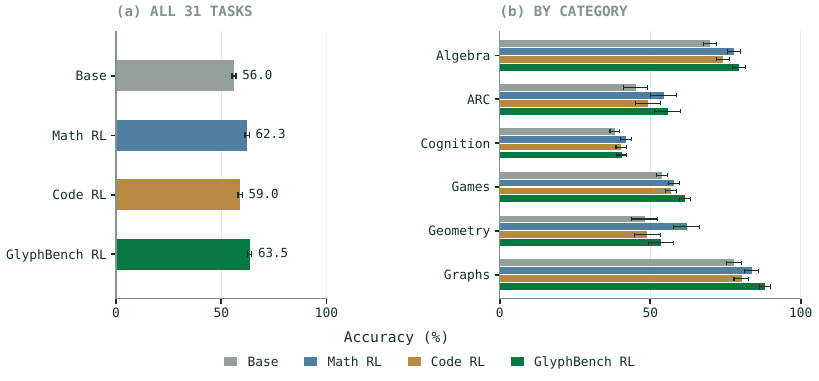}
  \caption{Reasoning Gym accuracy at a 32K output-token budget for the base model and policies trained on games, math, or code. Each policy answers the same 3,100 problems, with three attempts per problem. Panel (a) averages accuracy across 31 tasks; (b) separates the results by category. Whiskers show 95\% problem-bootstrap intervals, resampling within tasks and keeping each problem's three attempts together.}
  \label{fig:transfer}
\end{figure}

We now test whether the improvements acquired through gameplay extend to reasoning problems outside the training environments. We evaluate the base model and policies obtained after GlyphBench, Math, and Code RL on the same 3,100 Reasoning Gym problems with a 32K output-token budget (Figure~\ref{fig:transfer}). GlyphBench RL reaches 63.48\% accuracy, compared with 62.29\% for Math RL, 58.98\% for Code RL, and 56.04\% for the base model. The gain over Math RL is 1.19 percentage points, with a paired 95\% interval of $[0.38,2.02]$.

The gains extend across several forms of reasoning. GlyphBench RL leads on algebra, ARC, games, and graphs, while Math RL leads on cognition and geometry. Training through gameplay can therefore improve performance on problems that require a direct answer, even though the policy learned from sequences of actions and environmental feedback. These results provide empirical evidence that gameplay can yield stronger reasoning transfer than math or code RL in this setting.

In Appendices~\ref{app:transfer} and~\ref{app:baseline-transfer}, we report evaluations at additional output budgets, category results, and generation diagnostics. We also evaluate on MathArena and SWE-bench in Appendix~\ref{app:transfer}, where the changes are inconclusive.

\section{Conclusion}

Our experiments establish GlyphBench as a common setting for evaluating agents, studying learning dynamics, and post-training models whose gains can extend beyond gameplay. Our Craftax experiments show that glyph observations improve performance, while the benefit of memory and guidance depends on the model and its reasoning effort. Policies learn on individual tasks and across a task mixture, and the Snake experiments show how learning rate affects the benefit of frequent updates from smaller rollout groups. Gameplay training also improves performance on Reasoning Gym, outperforming math- and code-trained baselines in the main experiment.

The gains on Reasoning Gym suggest that learning through gameplay can contribute to capabilities that are useful beyond the games themselves. A next step is to understand which kinds of interaction support this transfer and how training can encourage it. GlyphBench makes it possible to study these questions across different tasks, connecting changes in performance to the observations and decisions made during learning.

\label{main-text-end}

% Keep references together after the main text in both paper formats.
\FloatBarrier
\clearpage
\subsection*{AI use statement}

We used LLMs to polish the writing and correct grammar.

\bibliography{references}

\clearpage
\appendix
\section{Extended Related Work}
\label{app:related}

\paragraph{Game benchmarks as experimental infrastructure.}
Over the past decade, the RL community has widely used game environments to develop and compare learning algorithms. The Arcade Learning Environment (ALE) helped establish this approach by bringing games with diverse control problems under a shared interface \citep{bellemare2013arcade}. Deep Q-learning subsequently demonstrated the value of this common testbed for studying a single learning algorithm across tasks \citep{mnih2015human}. A complementary approach is to construct smaller diagnostic problems. The Behaviour Suite separates properties such as exploration, memory, and robustness so that a result can be connected to a specific algorithmic strength or failure \citep{osband2020bsuite}. GlyphBench follows both traditions: its collection spans familiar game families, while focused tasks and replay help researchers investigate the behavior underlying aggregate performance. This breadth lets researchers ask where a method helps: for example, whether improved exploration in one game also benefits tasks that require memory or careful resource use.

\paragraph{Partial observability, generalization, and coordination.}
DeepMind Lab and ViZDoom provide configurable first-person environments for studying navigation, perception, and decisions from incomplete observations \citep{beattie2016lab,kempka2016vizdoom}. MiniGrid offers compact, configurable goal-directed tasks, and MiniHack uses the NetHack engine to vary navigation and the composition of game skills \citep{chevalierboisvert2023minigrid,samvelyan2021minihack}. Procgen explicitly separates training levels from unseen levels, making generalization part of evaluation \citep{cobbe2020leveraging}; Obstacle Tower combines procedural variation with visual control and planning \citep{juliani2019obstacle}. SMAC isolates a different challenge: decentralized cooperation when each unit acts from local observations \citep{samvelyan2019smac}. Together, these environments have made it possible to study capabilities that a single game would leave entangled or untested. An agent that excels at reactive control may still struggle to remember a location, generalize to an unseen level, or coordinate with another agent. GlyphBench brings several of these established single-agent task families into a common language-model interface, so researchers can examine how an intervention affects different demands on the policy.

\paragraph{Exploration and dependencies between goals.}
Crafter evaluates progress through achievements that require resource collection, survival, and crafting, connecting short decisions to longer sequences of prerequisites \citep{hafner2022crafter}. Craftax expands the mechanics and progression while using a fast JAX implementation to support large RL experiments \citep{matthews2024craftax}. NetHack adds a much larger space of interactions and difficult exploration problems \citep{kuttler2020nethack}. In Minecraft, MineRL pairs human demonstrations with a simulator to study sample-efficient learning, while MineDojo connects open-ended tasks to large collections of external knowledge \citep{guss2019minerl,fan2022minedojo}. The cost of studying these dependencies grows with the length and complexity of an episode. GlyphBench addresses this tradeoff with focused tasks for repeated optimization experiments at bounded interaction budgets, alongside extended environments in which agents pursue goals across longer sequences of decisions.

\paragraph{Text as an interface to a world.}
TextWorld allows researchers to generate text games with controlled difficulty and language, including variations in partial observability and reward sparsity \citep{cote2018textworld}. Jericho exposes human-authored interactive fiction, where agents face language understanding, commonsense reasoning, and large combinatorial action spaces \citep{hausknecht2020jericho}. These environments established text as an interface to worlds in which actions change what the agent can observe and do next. GlyphBench extends this use of text through observations that place objects in a two-dimensional grid and supply a legend and status display. By exposing adjacency and geometry directly while retaining a discrete action vocabulary, the interface makes spatial reasoning a central part of language-based interaction.

\paragraph{Using pretrained knowledge during interaction.}
ReAct interleaves reasoning with actions and feedback, providing a general pattern for adapting plans as observations arrive \citep{yao2023react}. In Crafter, SPRING organizes reasoning around game knowledge extracted from the environment's paper \citep{wu2023spring}; in Minecraft, Voyager maintains an expanding library of executable skills \citep{wang2023voyager}. ELLM uses a language model to propose meaningful goals that guide another agent's exploration, illustrating a different role for pretrained knowledge \citep{du2023ellm}. CODE-SHARP studies continual skill discovery and composition through hierarchical reward programs in Craftax \citep{bornemann2026codesharp}. Across these approaches, pretrained knowledge becomes useful through mechanisms that connect it to the current game state: plans, memory, skills, or exploration goals. GlyphBench's harness experiments examine this connection by varying the support available to a model during play, while trajectory inspection helps reveal how the model uses it.

\paragraph{From evaluating agents to training them.}
BALROG brings existing RL games to language and vision-language models and reports substantial weaknesses in long-horizon decision-making and visual grounding \citep{paglieri2024balrog}. Agentick evaluates multiple agent paradigms on purpose-built tasks with several observation modalities, finding large effects of reasoning harnesses and observation format \citep{castanyer2026agentick}. lmgame-Bench studies prompt sensitivity, perception, and memory in a shared game interface \citep{hu2025lmgame}. On the learning side, GLAM updates language-model policies through online environment interaction \citep{carta2023glam}. RAGEN examines training instability and the effects of rollout construction, and AgentGym-RL provides a modular framework for multi-turn RL across diverse interactive tasks \citep{wang2025ragen,xi2025agentgymrl}. GlyphBench builds on this combined evaluation and training agenda: a researcher can hold the task interface fixed while changing observations, harnesses, optimization settings, or training mixtures, then inspect the resulting policy behavior.

\paragraph{Transfer beyond the training environment.}
Automatically verifiable mathematics and coding tasks provide direct outcome feedback for language-model learning \citep{hendrycks2021math,chen2021humaneval,guo2025deepseek}. Reasoning Gym broadens this approach through procedurally generated reasoning tasks and verifiers \citep{stojanovski2025reasoninggym}. Games raise a related question: can rewards from sequential interaction improve performance when the model later faces a different kind of problem? The initial lmgame-Bench study reports transfer from single-game RL to unseen games and external planning tasks \citep{hu2025lmgame}; ViGaL finds that visual-game RL improves performance on multimodal mathematical and spatial reasoning evaluations \citep{xie2025vigal}. We study this question by training a shared policy on GlyphBench and comparing its transfer with a math-trained baseline and a code-trained baseline. Held-out game seeds measure learning on new instances, while external reasoning evaluations test capabilities outside the training environment. This combination supports broader research on how language models learn from rewards and generalize across problem formats.

\clearpage
\section{Environment and Interface Details}
\label{app:environment}

Each \textbf{GlyphBench} task specifies the actions, rewards, termination conditions, and observations in the decision process of Section~\ref{sec:contract}. The harness determines which observations, history, and memory the model receives. Standard tasks cap episodes at 512 steps and bound cumulative returns to $[-1,1]$. The same seed and action sequence reproduce an episode, including its stochastic events.

\begin{table}[!htb]
  \caption{The \NumAllTasks{} GlyphBench tasks. Standard tasks bound episodes to 512 steps and returns to $[-1,1]$. Extended tasks support longer episodes and are excluded from the standard aggregate score.}
  \label{tab:suites}
  \centering
  \small
  \setlength{\tabcolsep}{4pt}
  \begin{tabular}{lrlp{2.55in}}
    \toprule
    Suite & Tasks & Track & Task scope \\
    \midrule
    Classics & 50 & Standard & Puzzles and familiar games for planning, memory, and control. \\
    MiniGrid & 71 & Standard & Navigation, keys, doors, obstacles, and memory. \\
    MiniHack & 63 & Standard & Dungeon navigation, items, combat, and skill composition. \\
    Craftax & 60 & Standard & Focused survival, crafting, combat, and progression tasks. \\
    MiniAtari & 43 & Standard & Arcade control with compact dynamics and short horizons. \\
    Procgen & 16 & Standard & Procedural platformers, shooters, and mazes. \\
    \midrule
    Atari & 57 & Extended & Long-horizon arcade adaptations. \\
    Full Craftax & 1 & Extended & Full-game survival and progression. \\
    NetHack & 1 & Extended & Full-game exploration and progression. \\
    \bottomrule
  \end{tabular}
\end{table}

\subsection{Training interfaces and model actions}

Tasks follow Gymnasium's interaction convention \citep{towers2024gymnasium}: an episode begins with an initial observation, and each action produces a new observation and reward. The common interface lets researchers apply the same training and evaluation procedure across tasks with different rules and objectives.

A Verifiers adapter presents episodes as multi-turn model interactions with reward scoring \citep{brown_verifiers_2025}. The model chooses named actions from the task's vocabulary. Invalid responses consume a turn without advancing the agent's goal. Harnesses can maintain memory through separate model calls; the multitask study in Appendix~\ref{app:multitask} uses conversation history without a separate memory store.

\subsection{Inspecting an episode through replay}

\begin{figure}[!htbp]
  \centering
  \includegraphics[width=\linewidth]{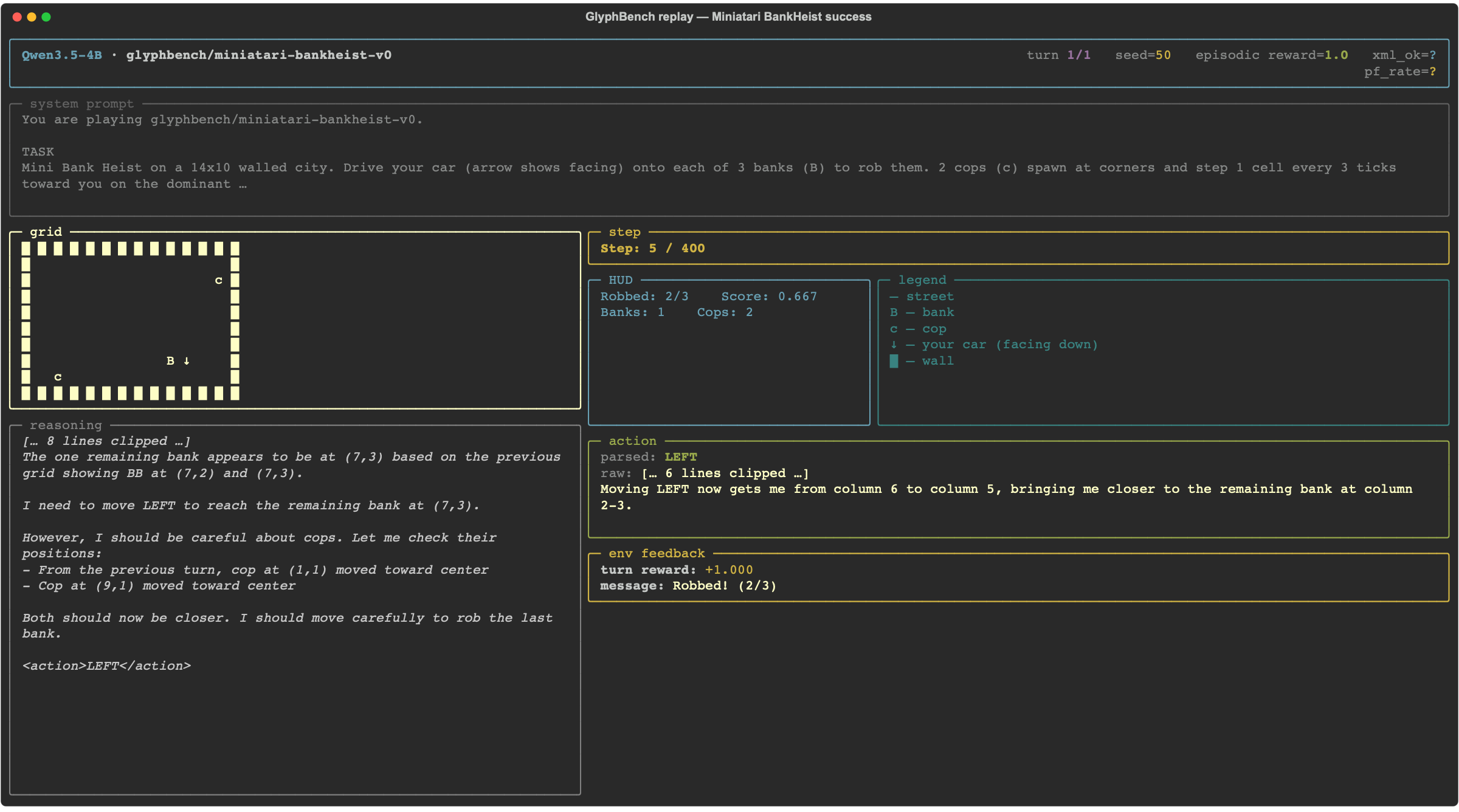}
  \caption{The final decision of a successful MiniAtari BankHeist episode. Replay connects the agent's observation and reasoning to its chosen action and the resulting environment feedback.}
  \label{fig:replay-ui}
\end{figure}

We provide a replay UI that aligns observations, generated reasoning, actions, rewards, and optional memory. In Figure~\ref{fig:replay-ui}, we show the final decision of a successful MiniAtari BankHeist episode. Viewing the observation alongside the agent's reasoning and action makes it possible to examine what information the agent used and how the environment responded.

Researchers can inspect the decisions behind a learning curve: whether an agent navigates more directly, conserves resources, recovers from a mistake, or repeats actions that leave it stuck. Connecting model outputs to their consequences helps turn changes in aggregate return into hypotheses about how the policy learns and interacts.

\FloatBarrier

\section{Craftax Observations and Agent Harnesses}
\label{app:craftax-interface}

\begin{figure}[!htb]
  \centering
  \includegraphics[width=0.95\linewidth]{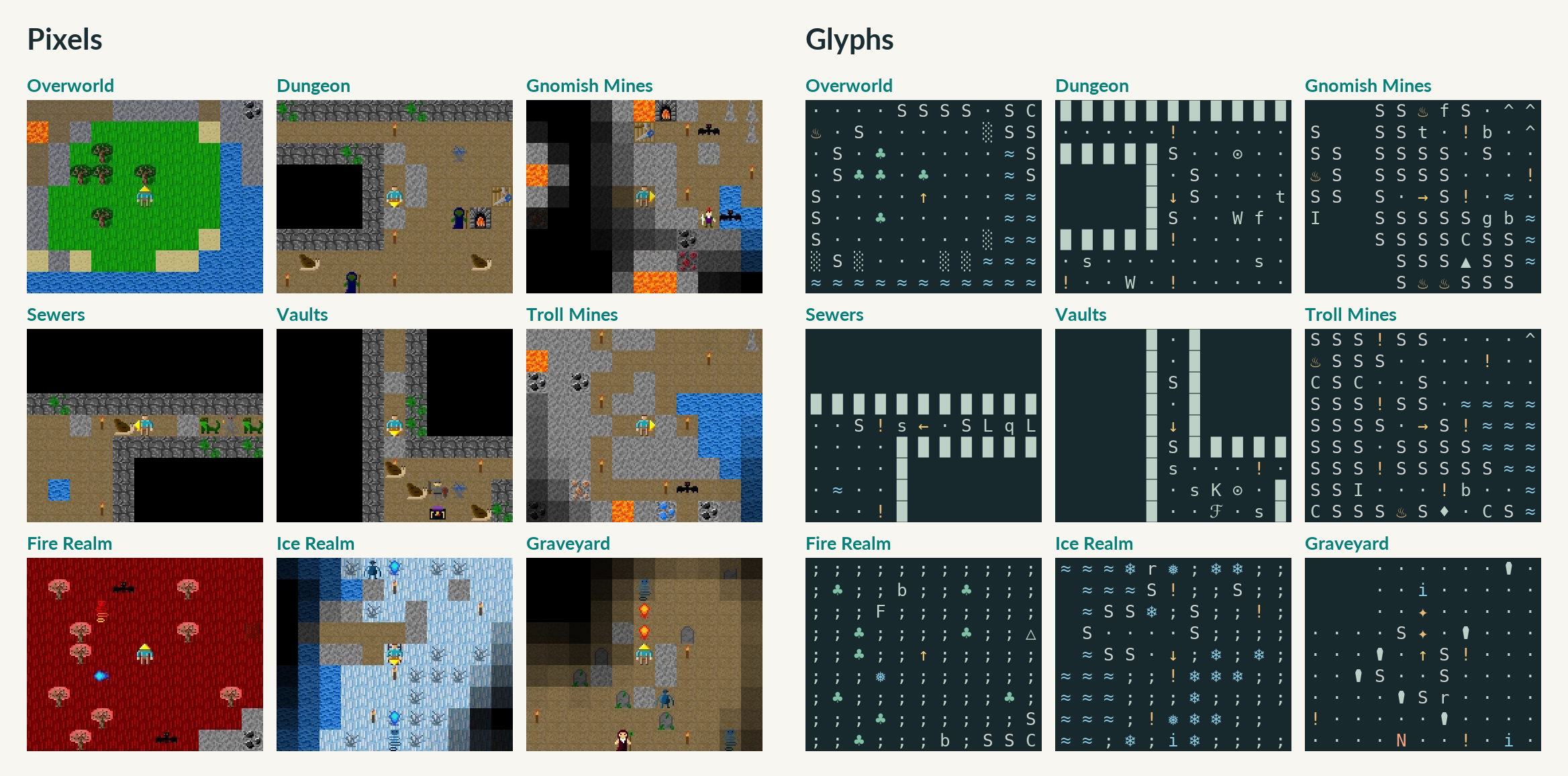}
  \caption{The nine Craftax floors in pixels and glyphs. Each pair shows the same local state, with HUDs and legends omitted. Blank glyph regions are unobserved cells; colors aid visualization only. The paired views show how the glyph interface represents the terrain and entities encountered during progression.}
  \label{fig:craftax-interface-floors}
\end{figure}

Our full Craftax wrapper exposes the original nine-floor game through the same action interface as the standard tasks. Progress requires agents to connect decisions across floors: they must find resources, remember useful locations, and revisit them as their equipment and goals change. We use this setting to study how observations and memory help agents plan, track progress, and adapt as an episode unfolds. Figure~\ref{fig:craftax-interface-modalities} compares the three observation formats, and Figure~\ref{fig:craftax-interface-floors} shows how the glyph interface represents each floor. The following sections describe the interfaces and harnesses used in Section~\ref{sec:observations}.

\subsection{Three observation formats}

\paragraph{Pixels.}
The native renderer supplies a local view with sprites, lighting, and an inventory/status panel. From this image, the model must identify terrain and entities, determine which way the player faces, and read equipment and resource levels. Only the current view is visible.

\paragraph{Native text (original Craftax renderer).}
We use Craftax's original text renderer as a comparison interface; it is not a GlyphBench observation format. It lists each cell in the local $9\times11$ window using coordinates relative to the player: $(x,y)=(0,0)$ is the current cell, positive $x$ points right, and positive $y$ points down. A cell can describe an actor or item together with its underlying terrain. Cells outside the light mask are reported as \texttt{Darkness}. Separate fields report inventory, equipment, vitals, attributes, facing, floor, and status flags.

\paragraph{Glyphs.}
The glyph renderer places one character at each location in the same $9\times11$ window, preserving the spatial arrangement of visible objects. A directional player glyph shows facing, and a legend identifies the visible symbols. The accompanying HUD reports vitals, inventory, equipment, floor, and absolute row/column position. Like native text, the renderer respects the local light mask. The compact representation merges some distinctions: grass and path share a glyph, for example, and an actor occupies the character that would otherwise show the terrain beneath it.

These design choices offer different ways to recover the game state. Native text explicitly names terrain and overlapping objects; glyphs make adjacency and relative position visible in the grid and supply absolute position in the HUD. Pixels convey appearance and continuous lighting, while both text formats apply a visibility threshold. Together, the interfaces let us examine how agents use spatial structure and explicit state descriptions during play.

\begin{figure}[p]
  \centering
  \includegraphics[width=\linewidth]{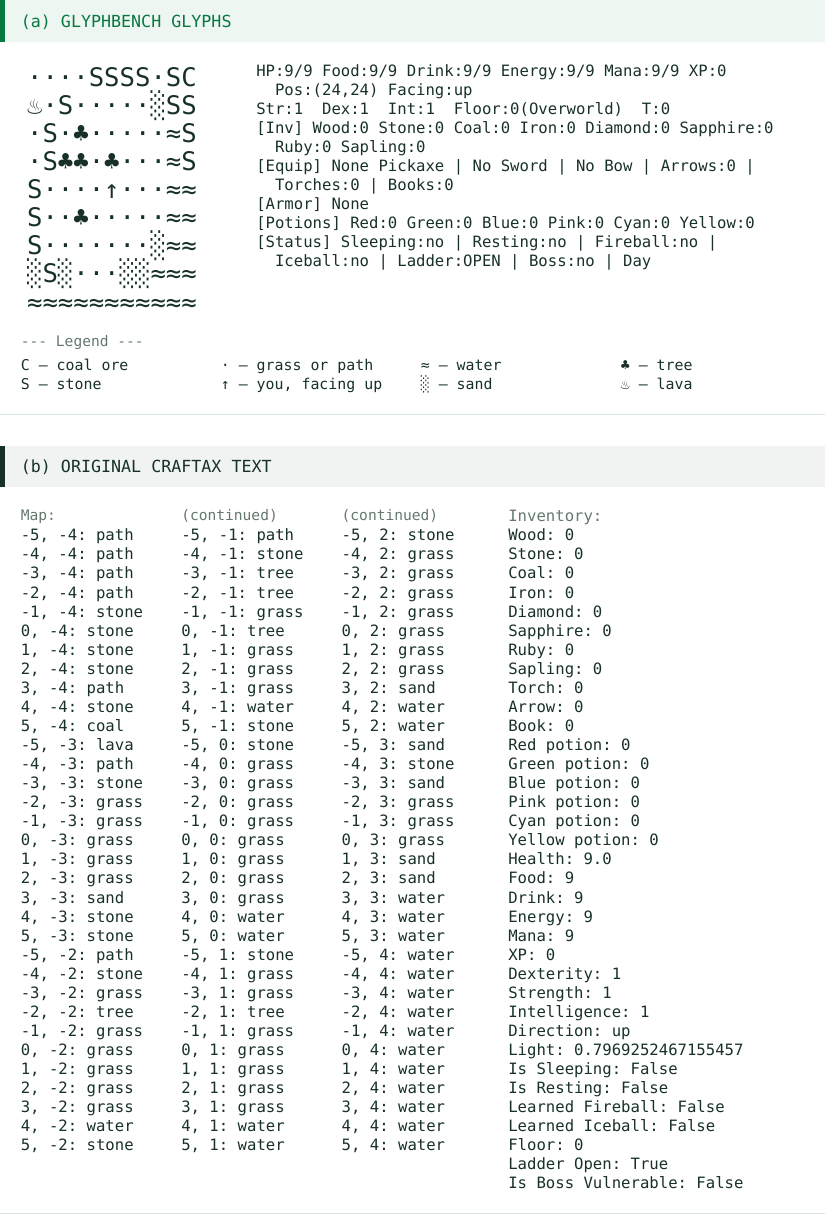}
  \caption{Complete environment observations for the initial Craftax state shown in Figure~\ref{fig:craftax-interface-modalities}. (a) GlyphBench's glyph grid, full HUD, and symbol legend. (b) Craftax's original text renderer: all 99 cells, inventory, and status information. The map listing continues down three columns. Line wrapping and column placement are adjusted for readability.}
  \label{fig:craftax-full-observations}
\end{figure}

\subsection{Basic and Guided Memory}
\label{app:craftax-harnesses}

We compare two agent harnesses to study how models use information accumulated during play. The \textit{Basic} harness combines standard game instructions with the growing conversation, allowing the model to consult previous observations and actions. The \textit{Guided Memory} harness organizes this information into a structured scratchpad, remembered landmarks, recent actions, and an exploration summary. It also supplies strategic advice for the current floor and uses separate model calls to update memory. Table~\ref{tab:craftax-harnesses} details the instructions and state available through each harness. Both choose one native game action per turn and cannot execute scripts. Memory updates store information without acting on the environment.

\begin{table}[!htbp]
  \centering\small
  \setlength{\tabcolsep}{5pt}
  \renewcommand{\arraystretch}{1.12}
  \caption{Basic and Guided Memory in the Craftax harness comparison. Both use glyph observations, native game actions, and the selected model reasoning-effort setting. Neither requests an explicit output-token cap.}
  \label{tab:craftax-harnesses}
  \begin{tabular}{@{}p{0.19\linewidth}p{0.34\linewidth}p{0.40\linewidth}@{}}
    \toprule
    {\raggedright Component\par} & {\raggedright Basic\par} & {\raggedright Guided Memory\par} \\
    \midrule
    {\raggedright Instructions\par} & {\raggedright Task, core mechanics, observation conventions, and action format\par} & {\raggedright Extended game guide, action format, memory protocol, and current-floor guidance\par} \\
    {\raggedright Current view\par} & {\raggedright Glyph grid and standard wrapper HUD\par} & {\raggedright Glyph grid and expanded inventory/status HUD\par} \\
    {\raggedright History\par} & {\raggedright Growing conversation, with the last eight observation/action frames rendered in each turn\par} & {\raggedright Last eight actions and rewards, plus persistent memory\par} \\
    {\raggedright External memory\par} & {\raggedright No separate memory store\par} & {\raggedright Six scratchpad sections and a floor-keyed landmark database\par} \\
    {\raggedright Exploration\par} & {\raggedright Prior observations in the conversation\par} & {\raggedright Summary of observed tiles, visited coordinates, frontiers, and remembered features\par} \\
    {\raggedright Memory calls\par} & {\raggedright None\par} & {\raggedright Every five steps and after relevant events; separate from action selection\par} \\
    \bottomrule
  \end{tabular}
\end{table}

The Guided Memory scratchpad records strategy, plans, lessons, tactical state, and floor notes to help agents revise their decisions as the game changes. Remembered landmarks and an exploration summary connect these plans to locations the agent has visited. The agent updates memory every five steps and after events that may require a revised plan, such as earning a reward, unlocking an achievement, or changing floors.

For the main comparisons, we evaluate each condition on five shared environment seeds using the model identifiers \texttt{gpt-5.6-luna}, \texttt{gpt-5.6-terra}, and \texttt{gpt-5.6-sol}. Both harnesses use the selected reasoning-effort setting without an additional output-token cap. Guided Memory also makes separate calls to maintain its memory, allowing us to examine whether models can use this additional support to improve progression.
\paragraph{Exploratory Astra run.}
We report one exploratory Astra run with glyph observations, Guided Memory, and maximum reasoning effort to illustrate how far a frontier model can progress with this combination. Astra reaches 51.4\% of maximum reward after 2,304 steps, shown by the star in Figure~\ref{fig:craftax-systems}.
\FloatBarrier

\paragraph{An action turn in each harness.}
For illustration, an agent returning to a crafting table on its right might output \texttt{<action>MOVE\_RIGHT</action>}. Basic chooses this action from the observation and conversation history. Guided Memory also consults its saved plan, landmarks, recent actions, and floor guidance. Either way, the environment executes one move and returns a new observation. A separate memory call can record the table's location for later decisions.

\subsection{Prompt excerpts}

The following excerpts show how the harnesses guide reasoning and maintain persistent state. We reproduce the instructions verbatim, with line wrapping adjusted for typesetting. The full prompts also describe game mechanics, available actions, and observation conventions.

\paragraph{Basic: reasoning and action format.}
\begin{quote}\small\raggedright
``No explicit output-token cap is imposed: use the model's native reasoning budget and take as much reasoning as is useful.''

``After reasoning, end your response with exactly one final action tag on its own line.''
\end{quote}

\paragraph{Guided Memory: separating action selection from memory.}
\begin{quote}\small\raggedright
``Discuss candidate actions as plain text while reasoning; only the final \texttt{<action>...</action>} tag is executed. Do NOT update memory on this turn.''

``Do all of your strategic bookkeeping here: update the plan, record lessons, note what you explored, and maintain the landmark map.''

``GROUNDING: the [Observation] is always ground truth. Read it before trusting memory; memory may be stale. Use [Recent Actions] to detect loops --- if an action keeps failing or you are oscillating, change approach.''
\end{quote}

\paragraph{Guided Memory: an example of floor-specific guidance.}
\begin{quote}\small\raggedright
``Progress promptly: once you have a stone pickaxe, stone sword, \texttt{>=4} spare wood, and safe food/drink/energy, locating the open ladder and descending is the main objective. Coal/iron upgrades are useful but OPTIONAL on floor 0; do not overfarm common stone or delay indefinitely for iron armour because floor 2 is far richer in ore.''
\end{quote}

The following excerpt from the example memory update in the harness instructions shows how the agent records its plan and a landmark. It is an instructional example, not an evaluated episode.
\begin{quote}\small
\begin{verbatim}
PLAN:
1. [x] Place crafting table
2. [ ] Mine 3 stone for a furnace
MAP_ADD: area=0, type=water, row=20, col=30, note=large lake
MAP_ADD: area=0, type=crafting_table, row=24, col=24, note=mine
\end{verbatim}
\end{quote}

The floor-specific guidance helps the agent decide when preparation is sufficient and further resource collection would delay progress. The agent can combine this guidance with its memory of earlier decisions to choose its next objective.

\FloatBarrier

\clearpage
\section{BALROG Observation Interfaces}
\label{app:balrog-interfaces}

BALROG presents game states through environment-specific language interfaces \citep{paglieri2024balrog}. Table~\ref{tab:balrog-interfaces} summarizes the original formats used in our comparison. Our glyph variant presents spatial information through two-dimensional grids of symbols, following GlyphBench's observation design. The position of each symbol directly expresses its location relative to the surrounding cells.

\begin{table}[!htbp]
  \centering\small
  \setlength{\tabcolsep}{5pt}
  \renewcommand{\arraystretch}{1.15}
  \caption{Original BALROG observation formats for the four environments in the interface comparison. These are BALROG's environment-specific representations, distinct from GlyphBench's shared glyph format.}
  \label{tab:balrog-interfaces}
  \begin{tabular}{@{}p{0.16\linewidth}p{0.79\linewidth}@{}}
    \toprule
    Environment & Original BALROG language interface \\
    \midrule
    BabyAI & Descriptions of visible objects and their positions relative to the agent, including door state and carried objects, together with the language task. \\
    Baba Is AI & Active rules and descriptions of objects and movable rule-word blocks using horizontal and vertical offsets from the controlled object. \\
    Crafter & Descriptions of nearby terrain and entities by distance and direction, the object directly ahead, survival statistics, and inventory. \\
    MiniHack & Language-wrapper descriptions of surroundings, inventory, and player statistics, supplemented by a two-dimensional ASCII map. \\
    \bottomrule
  \end{tabular}
\end{table}

\begin{figure}[!htb]
  \centering
  \includegraphics[width=\linewidth]{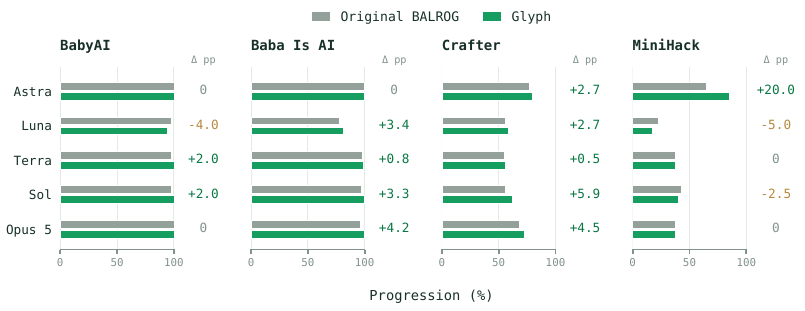}
  \caption{We compare the original BALROG interfaces with glyph observations on four environments. Bars show mean progression over paired episodes; differences are percentage points. Table~\ref{tab:balrog-interfaces} describes each original interface, including MiniHack's existing ASCII map.}
  \label{fig:balrog-interface}
\end{figure}

For BabyAI, Baba Is AI, and Crafter, glyph observations make spatial relationships explicit in the layout of the grid. MiniHack already supplies an ASCII map, so its comparison tests an alternative symbolic rendering. Figure~\ref{fig:balrog-interface} reports the resulting progression scores for each model and environment.

Glyph rendering increases 12 of 20 model-environment means, leaves five unchanged, and lowers three. All five models improve in mean progression on Crafter, while the largest increase occurs for Astra on MiniHack, from 65\% to 85\%. These comparisons show that the effect of the observation interface depends on both the environment and the model.
\FloatBarrier

\clearpage
\section{Training Objectives and Settings}
\label{app:training}

We start all training experiments from \BaselineModel{} and use AdamW with $(\beta_1,\beta_2)=(0.9,0.95)$, zero weight decay, gradient-norm clipping at 1, and a constant learning rate. Table~\ref{tab:training-settings} summarizes the settings used to establish single-task learnability, study optimization in Snake, and train on the multitask mixture.

\begin{table}[!htbp]
  \centering\small
  \caption{Training settings. Output budgets apply per action-generation call; sequence length is the trainer's token limit. Snake learning rates and group sizes are swept as described in Section~\ref{sec:protocol}.}
  \label{tab:training-settings}
  \begin{tabular}{lccc}
    \toprule
    Setting & Single-task runs & Snake & Multitask \\
    \midrule
    Learning rate & $10^{-6}$ & Sweep & $5\times10^{-6}$ \\
    Action output tokens & 8,192 & 4,096 & 4,096 \\
    Trainer sequence length & 32,768 & 32,768 & 65,536 \\
    Trainer loss & \texttt{default} & \texttt{ipo} & \texttt{ipo} \\
    \bottomrule
  \end{tabular}
\end{table}

\subsection{Group-relative advantages and policy updates}

Snake and multitask training use Prime-RL v0.9.0's group-relative estimator and the loss named \texttt{ipo} in that implementation \citep{prime2026rl}. Within the eligible trajectories of a rollout group $\mathcal G$, the advantage is
\begin{equation}
 A_i=R_i-\frac{1}{|\mathcal G|}\sum_{j\in\mathcal G}R_j.
 \label{eq:advantage}
\end{equation}
We assign this scalar to each trainable sampled token in trajectory $i$, so actions and reasoning tokens share the trajectory-level learning signal. Returns are mean-centered without standard-deviation normalization or a length penalty.

For sampled token $y_{ik}$ with model context $h_{ik}$, let $p_{ik}=\pi_\theta(y_{ik}\mid h_{ik})$ and let $q_{ik}$ be its recorded rollout-policy probability. Define $\rho_{ik}=p_{ik}/q_{ik}$ and let $m_{ik}$ select trainable RL tokens. The loss is
\begin{equation}
 \mathcal L=\frac{1}{N}\sum_{i,k}m_{ik}
 \left[-\mathbf{1}\{|p_{ik}-q_{ik}|\leq0.1\}\,A_i\rho_{ik}
       +10^{-3}(\log\rho_{ik})^2\right],
 \label{eq:training-loss}
\end{equation}
where $N=\sum_{i,k}m_{ik}$ counts trainable RL tokens across the batch. To limit the contribution of tokens whose probabilities have changed substantially, the probability-difference mask applies to the policy-gradient term; all trainable tokens still contribute to the penalty and denominator. The penalty measures change from the rollout policy. For the single-task learnability experiments, we use the trainer's \texttt{default} loss, with \texttt{dppo\_mask\_low} and \texttt{dppo\_mask\_high} both 0.2, $\tau_{\mathrm{adv}}=1$, and $\tau_{\mathrm{KL}}=0.001$.

\subsection{Task sampling and interaction history}

In Snake and multitask training, each rollout group contains trajectories from the same task. We cycle through the selected tasks and vary the game instances sampled within each group, giving the policy repeated exposure to each task's rules while changing the situations it encounters. Held-out evaluation tests new instances of those tasks.

During these runs, the policy conditions on conversation history and generates actions without separate memory-update calls. We sample at temperature 1 and treat invalid actions as no-ops. To study how best to allocate a fixed rollout budget, the Snake experiments vary group size and the resulting number of updates together. The multitask run instead uses 32 groups per update to combine learning signals from different tasks. We use these complementary settings to examine interactions between the learning rate, rollout grouping, and policy-update frequency.

\clearpage
\section{Single-Task Learning: Full Results}
\label{app:single-task}

\begin{figure}[!htbp]
  \centering
  \includegraphics[width=0.97\linewidth]{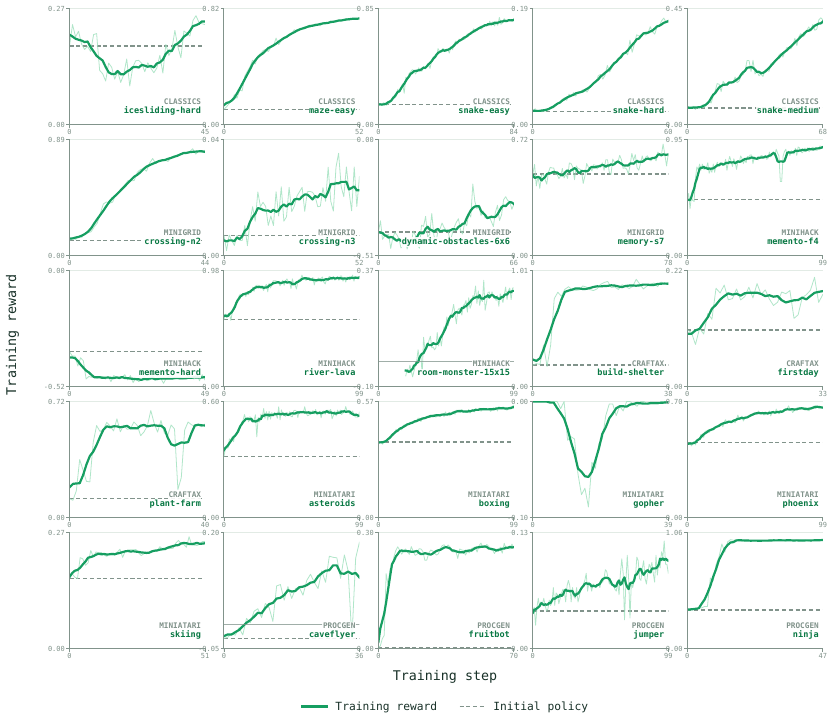}
  \caption{All \NumSingleTaskRuns{} single-task learning curves, ordered by suite and task. Pale traces show training rewards and dark traces show centered seven-point means. Dashed lines show initial-policy rollout means where available. Endpoint ticks indicate each task's training-step and reward range. All runs start from \BaselineModel{} and use one training seed.}
  \label{fig:single-full}
\end{figure}

We train separate policies across all six standard suites to check that the shared interface and task rewards support learning. In Figure~\ref{fig:single-full}, we show all \NumSingleTaskRuns{} learning curves. For the main-text figure, we select the alphabetically first and last task within each suite, giving 12 examples chosen independently of their learning outcomes.

The curves reveal both rapid improvement and slower, noisier learning. Because the vertical scales are task-specific, their heights should not be used to rank tasks or suites. Pale lines show the batch-level variation, while centered seven-point means make the overall trajectory easier to follow. Each curve measures training reward from one run. Together, they illustrate the range of learning dynamics that researchers can study with a common training pipeline.

\FloatBarrier

\clearpage
\section{Multitask Training and Evaluation}
\label{app:multitask}

\subsection{Task mixture and optimization}

We train a shared policy on a mixture of 100 tasks: 18 each from Classics, Craftax, MiniAtari, and MiniGrid, 19 from MiniHack, and nine from Procgen. Evaluation uses the same 100 task IDs with held-out seeds, measuring generalization to new instances of the training tasks.

Each update contains 1,024 rollouts in groups of 32. We use AdamW with learning rate $5\!\times\!10^{-6}$, $(\beta_1,\beta_2)=(0.9,0.95)$, gradient clipping at 1, no weight decay, and a constant schedule. The IPO-style loss uses $\tau_{\mathrm{KL}}=0.001$, $\tau_{\mathrm{adv}}=1$, and threshold 0.1. Appendix~\ref{app:training} defines the loss and task sampling. The sequence length is 65,536, with an output budget of 4,096 tokens per turn and no separate persistent memory store.

\subsection{Evaluation and learning curves}

We evaluate every 50 gradient steps on 200 examples drawn from four held-out seeds and weight the 100 task returns equally in the aggregate score. Here we show the full learning curves and use only observations within each displayed interval for smoothing.

\begin{figure}[!htbp]
  \centering
  \includegraphics[width=\linewidth]{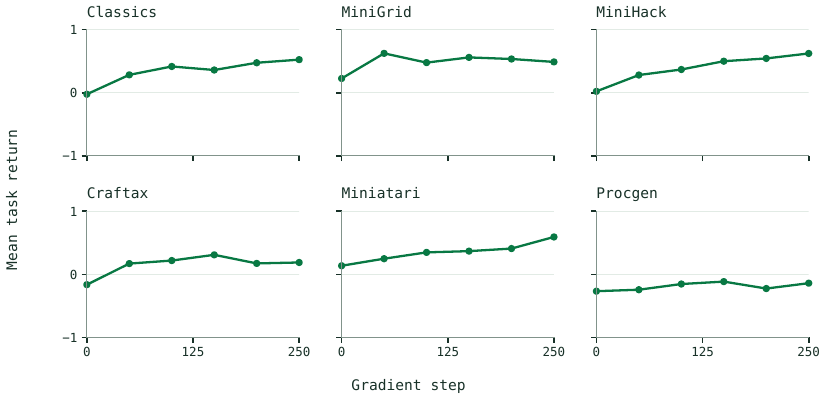}
  \caption{We evaluate the multitask policy on new environment seeds across all six training suites, using a common return scale. Each point averages task means within a suite. All suites improve over the displayed interval, while Procgen remains below zero.}
  \label{fig:multitask-suites}
\end{figure}

\begin{figure}[p]
  \centering
  \includegraphics[width=\linewidth]{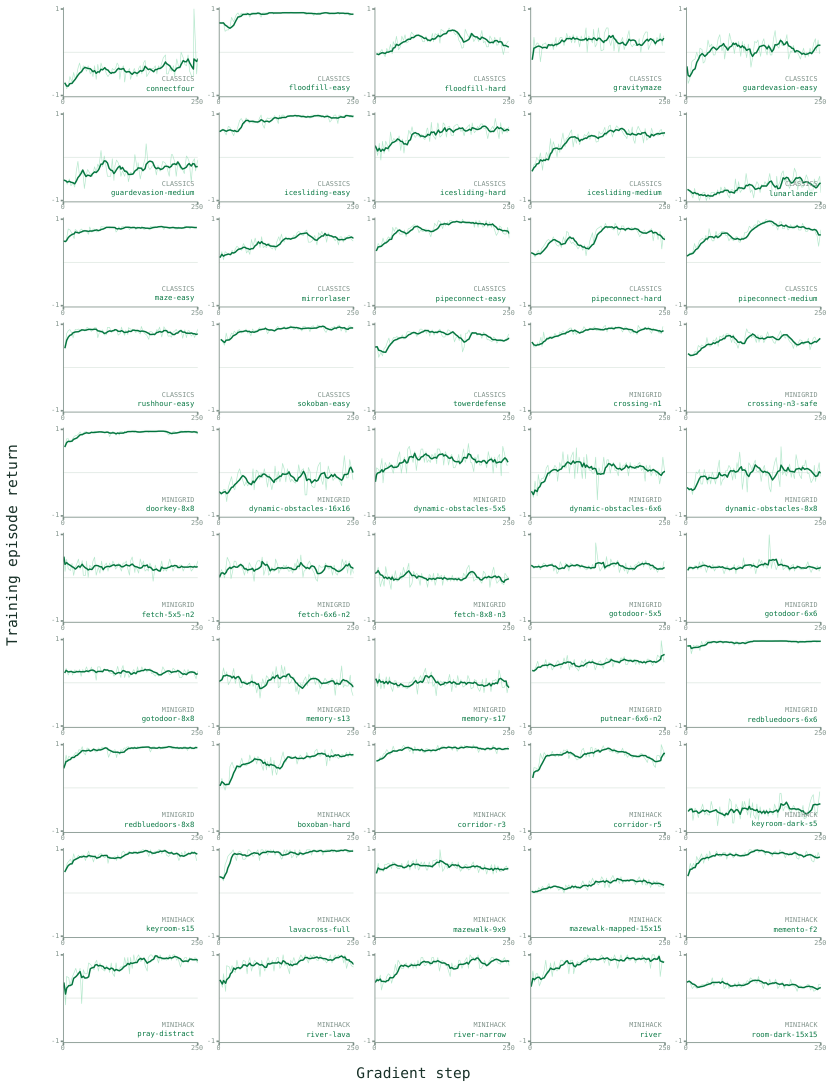}
  \caption{The multitask policy learns jointly on all 100 tasks; here we show tasks 1--50, ordered by suite and task. Pale lines show task-mean batch returns; dark lines show trailing 15-gradient-step means. All panels share steps 0--\HeroStep{} and returns $[-1,1]$.}
  \label{fig:hero-task-grid}
\end{figure}

\begin{figure}[p]
  \centering
  \includegraphics[width=\linewidth]{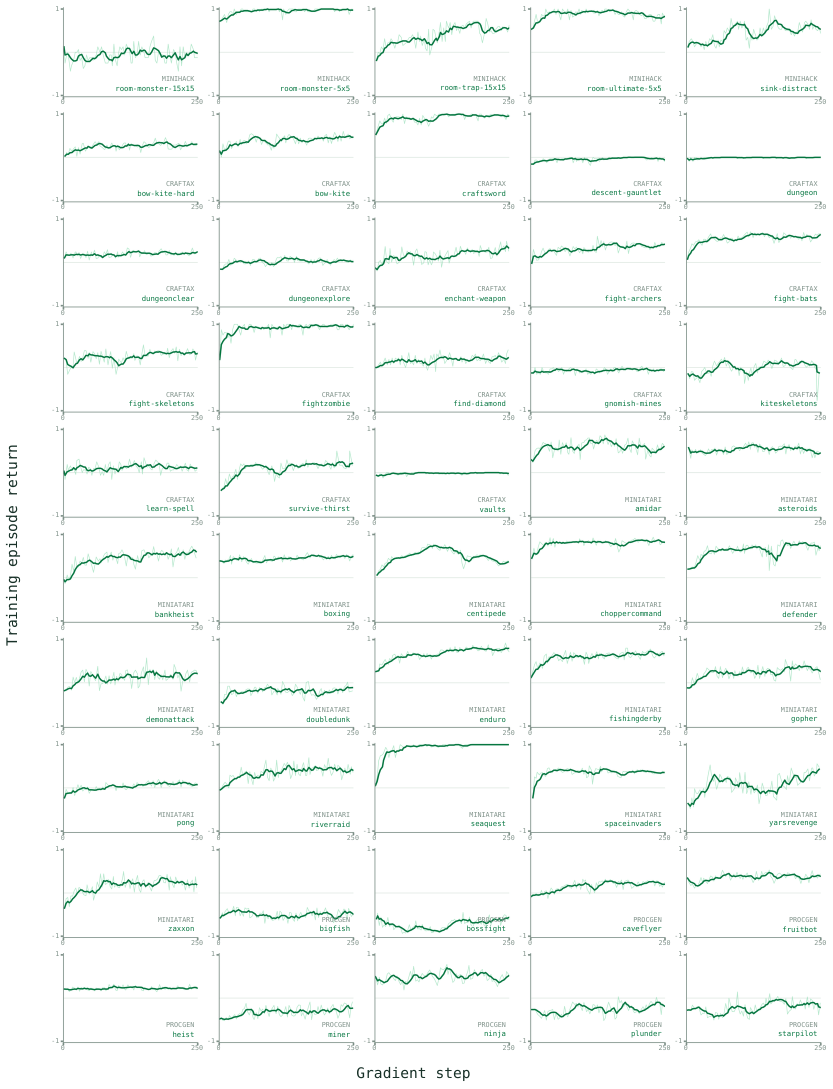}
  \caption{Training curves for the remaining 50 tasks of the same multitask policy. Axes and smoothing match Figure~\ref{fig:hero-task-grid}; together, the figures show each task once.}
  \label{fig:hero-task-grid-second}
\end{figure}

In Figure~\ref{fig:multitask-suites}, we show how held-out performance develops in each suite. We then plot all 100 task-level training curves in Figures~\ref{fig:hero-task-grid}--\ref{fig:hero-task-grid-second} to examine differences in the pace and stability of learning. These curves track the same shared policy throughout training.

\FloatBarrier

\clearpage
\clearpage
\clearpage

\section{External Transfer: Protocols and Full Results}
\label{app:transfer}

To measure how game training affects performance in other domains, we compare the base \BaselineModel{} and GlyphBench RL policy on reasoning, mathematics, and software-engineering benchmarks excluded from \textbf{GlyphBench} RL training.

\subsection{Sampling and inference budgets}

\paragraph{Reasoning Gym.}
We evaluate 31 tasks with 100 problems each and three completions per problem, totaling 3,100 problems and 9,300 attempts per model at each budget. Both models receive identical problems at output budgets of 2K, 8K, and 16K tokens. Solve rate averages success over all attempts, counting malformed or unparseable answers as failures.

\paragraph{Mathematics.}
MathArena evaluation covers seven AIME/HMMT datasets: 183 problems with four generations each, totaling 732 attempts per model. The output ceiling is 65,500 tokens, and accuracy is measured over all attempts without best-of-four selection. The 2K/8K/16K sweep applies only to Reasoning Gym.

\paragraph{Software engineering.}
We evaluate both policies on the same 500 SWE-bench Verified instances and 576 SWE-bench Pro instances, denoted Pro-576. Each episode allows at most 50 agent turns and 32,768 output tokens per model call. A patch succeeds if it passes the benchmark's tests.

\subsection{Paired uncertainty}

We compute paired bootstrap intervals from 10,000 resamples. Reasoning Gym resamples problems within tasks, keeping each problem's three attempts together. MathArena likewise groups each problem's four generations; SWE-bench resamples paired instances. These intervals capture evaluation uncertainty for one checkpoint, not variation across training seeds. Changes are computed from unrounded rates and may differ from the subtraction of displayed values.

\begin{figure}[!htbp]
  \centering
  \includegraphics[width=\linewidth]{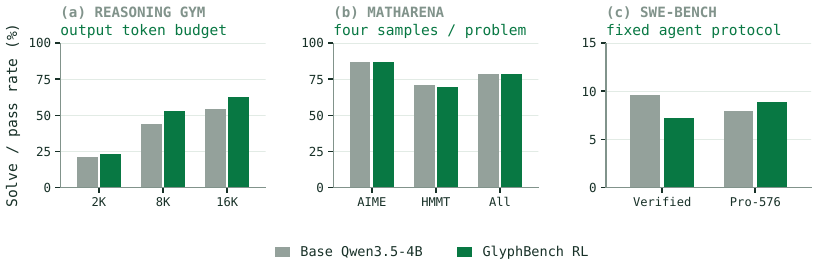}
  \caption{We evaluate transfer beyond gameplay by comparing performance before and after GlyphBench RL. (a) Reasoning Gym solve rates at three output budgets. (b) MathArena accuracy over all attempts. (c) SWE-bench pass rates. Bars start at zero; panel (c) spans 0--15\%. Table~\ref{tab:hero-transfer} reports paired changes and uncertainty.}
  \label{fig:transfer-cross-domain}
\end{figure}

\begin{table}[!htbp]
  \centering\small
  \setlength{\tabcolsep}{4pt}
  \caption{External evaluation of the base and GlyphBench RL models. Rates are percentages; changes and paired 95\% intervals are percentage points. $N$ counts attempts for mathematics and reasoning, and instances for coding. Resampling clusters attempts by problem.}
  \label{tab:hero-transfer}
  \begin{tabular}{lrrrrr}
\toprule
Benchmark & $N$ & Base & GlyphBench RL & $\Delta$ (pp) & Paired 95\% CI (pp) \\
\midrule
MathArena aggregate & 732 & 78.55 & 78.42 & $-0.14$ & $[-2.46, +2.19]$ \\
AIME 2024 I & 60 & 90.00 & 93.33 & $+3.33$ & $[+0.00, +10.00]$ \\
AIME 2024 II & 60 & 90.00 & 90.00 & $+0.00$ & $[-5.00, +5.00]$ \\
AIME 2025 & 120 & 85.00 & 85.00 & $+0.00$ & $[-5.00, +5.00]$ \\
AIME 2026 & 120 & 85.00 & 85.00 & $+0.00$ & $[-5.83, +5.83]$ \\
HMMT Feb 2025 & 120 & 70.00 & 64.17 & $-5.83$ & $[-14.17, +2.50]$ \\
HMMT Nov 2025 & 120 & 77.50 & 80.83 & $+3.33$ & $[-1.67, +8.33]$ \\
HMMT Feb 2026 & 132 & 65.15 & 65.15 & $+0.00$ & $[-4.55, +4.55]$ \\
SWE-bench Verified & 500 & 9.60 & 7.20 & $-2.40$ & $[-4.80, +0.00]$ \\
SWE-bench Pro-576 & 576 & 7.99 & 8.85 & $+0.87$ & $[-1.56, +3.30]$ \\
Reasoning Gym 2K & 9,300 & 21.22 & 23.40 & $+2.18$ & $[+1.51, +2.88]$ \\
Reasoning Gym 8K & 9,300 & 44.11 & 53.33 & $+9.23$ & $[+8.38, +10.09]$ \\
Reasoning Gym 16K & 9,300 & 54.23 & 62.58 & $+8.35$ & $[+7.43, +9.29]$ \\
\bottomrule
\end{tabular}

\end{table}

\subsection{Results across domains}

We also evaluate transfer on the 23 non-game Reasoning Gym tasks. At 8K, their solve rate improves from 48.52\% to 58.12\%, a gain of $9.59$ points with a 95\% interval of $[8.54,10.65]$. The improvement extends beyond the benchmark's game category. Tables~\ref{tab:hero-transfer}--\ref{tab:reasoning-categories} report aggregate and category results.

We find little aggregate change on MathArena, where a small AIME increase is offset by a small HMMT decrease. On SWE-bench, Verified declines in its point estimate and Pro-576 increases slightly; neither change establishes an improvement. We report the numerical breakdown and paired intervals in Table~\ref{tab:hero-transfer}. Exact two-sided McNemar tests give $p=0.0807$ for Verified and $p=0.5758$ for Pro-576. In these evaluations, the clearest transfer gains occur on Reasoning Gym.

\begin{figure}[!htbp]
  \centering
  \includegraphics[width=\linewidth]{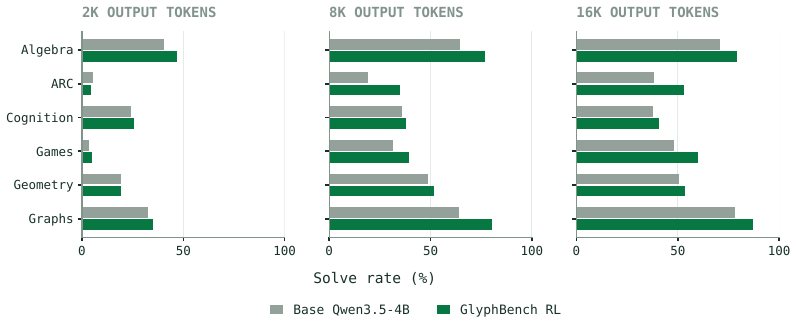}
  \caption{Reasoning Gym performance before and after GlyphBench RL by category and output budget, on a common 0--100\% scale. Every category improves at 8K and 16K; the smaller 2K budget yields a weaker and less consistent benefit. Table~\ref{tab:reasoning-categories} reports the corresponding values.}
  \label{fig:reasoning-categories}
\end{figure}

\begin{figure}[!htbp]
  \centering
  \includegraphics[width=\linewidth]{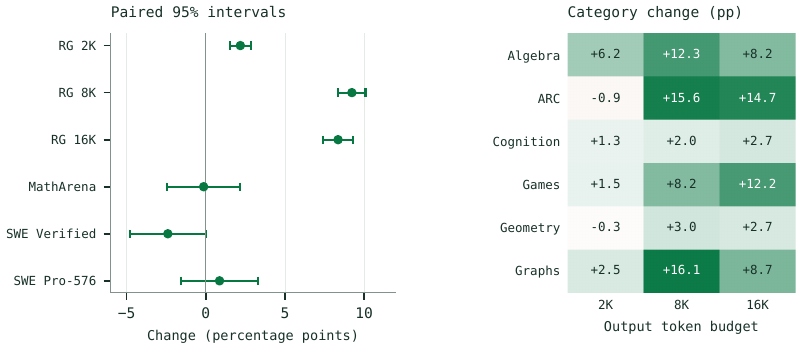}
  \caption{Changes after GlyphBench RL on external benchmarks, in percentage points. The left panel shows aggregate changes with paired 95\% bootstrap intervals; the right separates Reasoning Gym changes by category and output budget. Positive estimates whose intervals include zero do not establish an improvement.}
\end{figure}

\begin{table}[!htbp]
  \centering\small
  \setlength{\tabcolsep}{3pt}
  \caption{Reasoning Gym solve rates (\%): base / GlyphBench RL, with changes in percentage points. Categories contain 6, 3, 7, 8, 2, and 5 tasks in row order, each with 100 problems and three attempts per problem.}
  \label{tab:reasoning-categories}
  \begin{tabular}{lrrr}
\toprule
Category & 2K: base / ours ($\Delta$) & 8K: base / ours ($\Delta$) & 16K: base / ours ($\Delta$) \\
\midrule
Algebra & 40.61 / 46.78 ($+6.17$) & 64.72 / 77.06 ($+12.33$) & 71.00 / 79.22 ($+8.22$) \\
ARC & 5.22 / 4.33 ($-0.89$) & 19.22 / 34.78 ($+15.56$) & 38.22 / 52.89 ($+14.67$) \\
Cognition & 24.29 / 25.62 ($+1.33$) & 36.05 / 38.00 ($+1.95$) & 37.86 / 40.57 ($+2.71$) \\
Games & 3.38 / 4.92 ($+1.54$) & 31.42 / 39.58 ($+8.17$) & 47.88 / 60.08 ($+12.21$) \\
Geometry & 19.33 / 19.00 ($-0.33$) & 48.50 / 51.50 ($+3.00$) & 50.67 / 53.33 ($+2.67$) \\
Graphs & 32.53 / 35.00 ($+2.47$) & 64.13 / 80.20 ($+16.07$) & 78.20 / 86.93 ($+8.73$) \\
\bottomrule
\end{tabular}

\end{table}
\FloatBarrier

\section{Math and Code Baselines for Reasoning Transfer}
\label{app:baseline-transfer}

We compare GlyphBench RL with a math-trained baseline and a code-trained baseline to test how reasoning gains from gameplay transfer relative to these common RL domains. All policies start from the same \BaselineModel{} checkpoint. We evaluate the trained policies on identical Reasoning Gym problems using a shared generation and scoring procedure within each output budget.

\subsection{Training data and learning curves}

Math RL uses 100 fixed problems from DeepMath-103K \citep{deepmath2025}, restricted to high-difficulty questions with scalar numerical answers. We stratify the selection by topic and difficulty. Ambiguous, proof-based, and multipart formats are excluded; rewards use deterministic Math-Verify grading without an LLM judge, and teacher solutions are not supplied to the policy.

Code RL uses 100 fixed problems from the code portion of INTELLECT-3-RL \citep{intellect3rl2025}. Filtering removes labelled LiveCodeBench rows, duplicate questions, interactive tasks, unrendered visual dependencies, and tasks requiring arbitrary valid outputs. Sampling stratifies by historical solve rate, source, and prompt length. A response receives reward only if it passes every supplied test in the pinned execution environment.

\begin{figure}[!htbp]
  \centering
  \includegraphics[width=\linewidth]{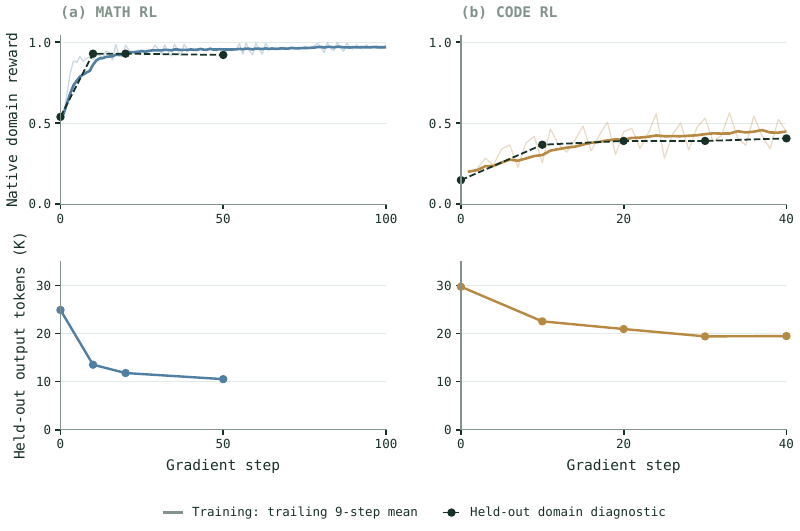}
  \caption{Math and code RL improve rewards in their respective training domains. The top panels show training reward (pale), trailing nine-step means (solid), and held-out domain reward (dashed). The bottom panels show average output tokens on held-out problems from each domain. These rewards measure learning within each domain; we evaluate reasoning transfer separately on Reasoning Gym.}
  \label{fig:baseline-learning}
\end{figure}

Both baselines use the group-relative advantages and IPO-style objective described in Appendix~\ref{app:training}, with learning rate $5\!\times\!10^{-6}$, groups of 32, and batches of 1,024 rollouts. Baseline training uses single-turn responses with a 32,768-token output allowance and 65,536-token context. Figure~\ref{fig:baseline-learning} shows the recorded learning curves and held-out diagnostics in each training domain.

The math policy rapidly improves its domain reward and then levels off, whereas code learning is slower and noisier. Both also reduce the average length of responses in their held-out domain diagnostics. Reward heights across the two domains are not directly comparable: the verifiers, problem distributions, and opportunities for improvement differ. We compare their reasoning capabilities on the shared external evaluation below.

\FloatBarrier
\subsection{Reasoning Gym protocol}

We use 31 tasks with 100 problems per task and three completions per problem. Each model receives the same 3,100 problems and produces 9,300 attempts at each cap. Sampling uses temperature 0.6, top-$p$ 0.9, the DeepSeekZero system prompt, and \texttt{<answer>} extraction.

Strict per-completion accuracy counts a native verifier score of at least 1 as success and divides by all attempts. Partial credit, malformed answers, and grading exceptions do not count as successes. Observed pass@3 instead counts a problem as solved if any of its three recorded completions succeeds. Overall accuracy weights tasks equally; because all tasks have the same number of attempts, this is also the mean over completions. It is not the unweighted mean of the six category means.

The main comparison uses a 32,768-token output cap with a 65,536-token model context. The supplementary 2K, 8K, and 16K evaluations use a 32,768-token context. Each cap is a separate sampling run, rather than a truncated prefix of the same responses. Within a cap, all four models use the same inference configuration.

\subsection{Accuracy, paired differences, and category results}

At 32K, GlyphBench RL achieves the highest accuracy. Relative to the base model, its gain is 7.44 percentage points; relative to Math RL and Code RL, the gains are 1.19 and 4.51 points. Table~\ref{tab:baseline-paired} reports paired uncertainty, computed by resampling problems within tasks and retaining each problem's three completions together. The 10,000 bootstrap resamples quantify evaluation uncertainty for the trained policies; variability across independent training runs requires repeated experiments.

\begin{table}[!htbp]
  \centering\small
  \caption{GlyphBench RL minus each comparison policy at the 32K output budget. Differences and paired 95\% bootstrap intervals are percentage points.}
  \label{tab:baseline-paired}
  \begin{tabular}{lrr}
    \toprule
    Reference policy & Accuracy difference & Paired 95\% interval \\
    \midrule
    Base & $+7.44$ & $[+6.55,+8.34]$ \\
    Math RL & $+1.19$ & $[+0.38,+2.02]$ \\
    Code RL & $+4.51$ & $[+3.62,+5.40]$ \\
    \bottomrule
  \end{tabular}
\end{table}

\begin{figure}[!htbp]
  \centering
  \includegraphics[width=\linewidth]{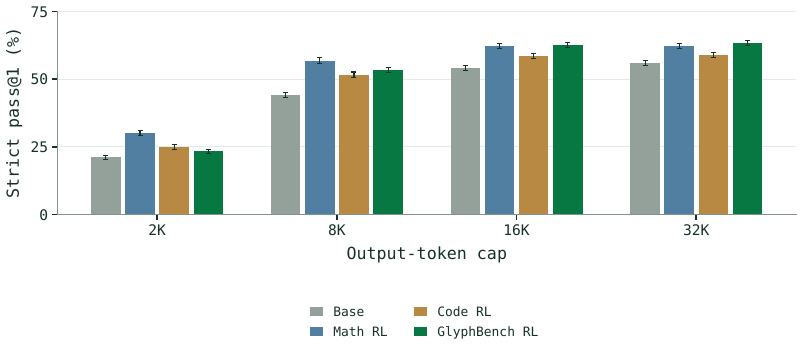}
  \caption{We compare Reasoning Gym accuracy at additional output-token budgets. Each cap uses independently sampled responses to the same problems. Whiskers show 95\% problem-bootstrap intervals; all three attempts for a problem remain in the same resampled cluster.}
  \label{fig:baseline-budgets}
\end{figure}

GlyphBench RL outperforms the math- and code-trained baselines on algebra, ARC, games, and graphs, while Math RL leads on cognition and geometry. In Figure~\ref{fig:baseline-budgets}, we compare the policies at additional output budgets. Math RL leads at 2K and 8K, and its difference from GlyphBench RL is inconclusive at 16K. These patterns suggest that the reasoning capabilities learned during RL interact with the compute available at evaluation.

\subsection{Response length and multiple attempts}

\begin{figure}[!htbp]
  \centering
  \includegraphics[width=\linewidth]{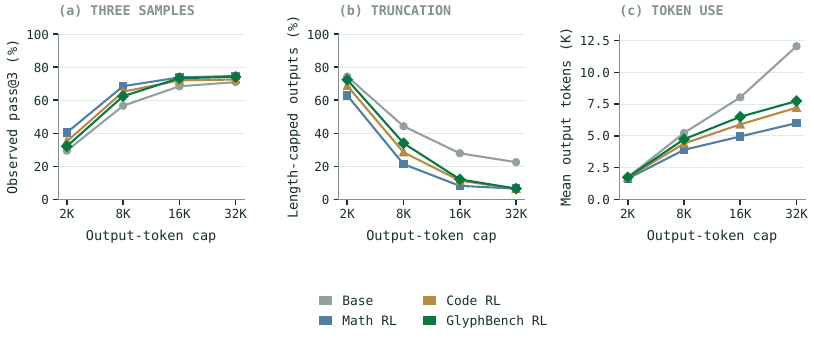}
  \caption{Reasoning Gym generation diagnostics. (a) Problems with at least one successful completion among three attempts. (b) Responses ending at the output-token limit. (c) Mean output tokens per completion. Truncated responses remain in the denominator and can still count as correct if the answer is extractable.}
  \label{fig:reasoning-diagnostics}
\end{figure}

Figure~\ref{fig:reasoning-diagnostics} separates three aspects of generation: whether any of three samples succeeds, how often generation reaches its length cap, and how many output tokens are used on average. At 32K, GlyphBench RL reduces truncation from 22.55\% to 6.54\% and mean output length from about 12.1K to 7.7K tokens relative to the base model. Higher accuracy therefore does not require longer average responses in this comparison. Math and code training also reduce truncation, so this behavior alone does not explain the difference between training domains.

The choice of metric matters as well. At 32K, Math RL has a slightly higher observed pass@3 than GlyphBench RL (74.74\% versus 74.19\%), although the paired interval for their difference includes zero. This measure uses the same three completions as the accuracy analysis and introduces no additional generation. A model can have higher average per-completion accuracy without solving more distinct problems across three tries.

\FloatBarrier

\clearpage
\section{Snake: Learning-Rate Stability}
\label{app:snake-collapse}

We use the Snake experiments to examine why higher learning rates can produce promising early improvements followed by collapse. Comparing each configuration's peak and final return shows which gains survive continued optimization.

The main Snake study covers 60 conditions: three difficulties, five group sizes, and four learning rates from $10^{-6}$ to $10^{-5}$. Figure~\ref{fig:snake} shows the 45 conditions at the three lower rates. We also evaluate six exploratory conditions at $10^{-4}$.

\begin{figure}[!htbp]
  \centering
  \includegraphics[width=\linewidth]{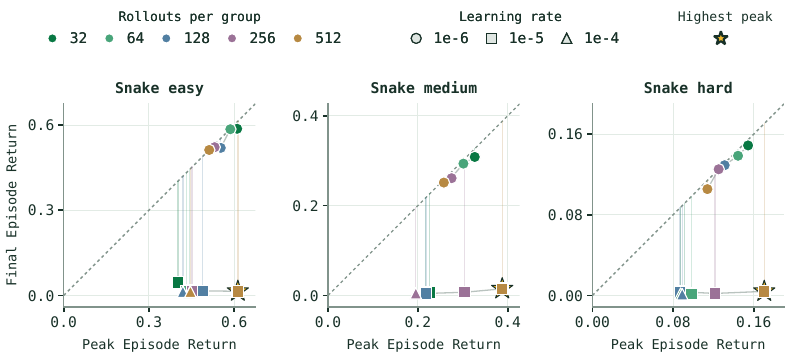}
  \caption{Peak versus final observed Snake return at $10^{-6}$, $10^{-5}$, and $10^{-4}$ within 6,144 rollouts. Both coordinates use trailing 512-rollout means. Colors indicate group sizes, shapes indicate learning rates, and gray paths connect increasing group sizes within a rate. The diagonal marks full retention of the peak; vertical stems show the drop to final return. Gold stars mark each difficulty's largest peak among plotted conditions. The $10^{-4}$ sweep is exploratory.}
  \label{fig:snake-collapse}
\end{figure}

Figure~\ref{fig:snake-collapse} compares peak and final observed returns within the 6,144-rollout budget, both measured from trailing 512-rollout means.

The stability comparison reveals different effects of group size across learning rates. At $10^{-6}$, larger groups generally reach lower peak and final returns. At $10^{-5}$, group 512 reaches the highest peak among the plotted conditions on every difficulty, yet ends near zero. On easy Snake, for example, return falls from a peak of 0.613 to 0.015. Selecting a configuration by peak return alone would obscure this instability: an apparent gain may not survive continued optimization. These collapses motivated the intermediate learning rates in Figure~\ref{fig:snake}.
\FloatBarrier

\clearpage
\section{Snake: Extending the Rollout Budget}
\label{app:snake-long}

To test whether the larger group catches up with more interaction, we compare groups 32 and 512 at learning rate $10^{-6}$ over 25,600 rollouts on all three difficulties. These six runs form a separate study from the 6,144-rollout sweep.

With this larger budget, group 32 achieves the higher final return on all three tasks. The gap is largest on easy Snake (0.824 versus 0.585) and remains positive as difficulty increases. Thus the small-group advantage at this learning rate persists beyond the early phase of learning. This comparison extends the fixed-interaction analysis; it still couples group size to the number of policy updates.

\begin{figure}[!htbp]
  \centering
  \includegraphics[width=\linewidth]{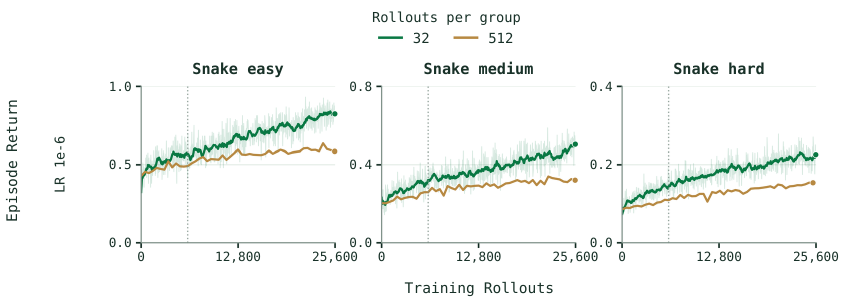}
  \caption{Snake training over 25,600 rollouts at $10^{-6}$, comparing groups 32 and 512 across three difficulties. Pale traces show batch returns, solid curves show trailing 512-rollout means, and dots mark final observations. The dotted line marks 6,144 rollouts for reference. Group 32 achieves the higher final return on all three tasks.}
  \label{fig:snake-long}
\end{figure}
\FloatBarrier

\clearpage
\section{Limitations and Scope of the Findings}
\label{sec:limitations}

\paragraph{The game abstraction.}
GlyphBench represents game states as spatial text and bounds the horizons of standard tasks to make language-model RL experiments practical. The grid preserves cell layout and named entities while abstracting away texture, perspective, and some visual ambiguity. Standard-task scores measure progress within these adaptations; evaluating mastery of a source game requires its native mechanics and horizons. Common return bounds make tasks easier to combine but do not equalize reward density, difficulty, or opportunities for improvement.

\paragraph{Agent-design comparisons.}
We compare observations, models, and harnesses as working agent configurations. Each observation format exposes different state information, and Guided Memory combines richer instructions with persistent memory and additional model calls. Focused ablations and fixed-budget comparisons would help separate these contributions to progression and cost. Similarly, the model comparisons evaluate whole systems without isolating parameter count or architecture.

\paragraph{Training studies.}
The experiments illustrate research questions that GlyphBench supports, including optimization stability, rollout allocation, and learning across tasks. Each condition uses one training run, so repeated experiments are needed to estimate the variability of the observed patterns. In Snake, fixing the rollout budget couples group size to update frequency; fixing update counts in a complementary study would help separate their effects. The multitask evaluation tests new instances of trained tasks, leaving transfer to entirely unseen GlyphBench tasks for future work.

\paragraph{Transfer.}
Our results show that reasoning gains from gameplay can transfer to held-out Reasoning Gym problems and exceed those of the math- and code-trained baselines in the main accuracy comparison. This evidence concerns the evaluated policies and protocol. The tasks may share skills with gameplay or overlap with model pretraining, and the bootstrap intervals quantify evaluation uncertainty rather than variation across training runs. Response-length diagnostics show less truncation and shorter average responses after training but do not identify the cause of improved accuracy. The policy ranking also changes with the output budget and metric: the accuracy leader need not lead on observed pass@3. We do not find comparably clear gains on the mathematics and software-engineering evaluations.

\end{document}